%% file: main.tex
\documentclass[11pt]{elsarticle}
\usepackage{booktabs}
\usepackage{epsfig}
\usepackage{amssymb}
\usepackage{amsmath}
\usepackage{amsfonts}
\usepackage{float}
\usepackage{adjustbox}
\def\*#1{\mathbf{#1}}
\usepackage{url}
\usepackage{booktabs}
\usepackage{cleveref}
\usepackage{soul}
\usepackage{color}
\usepackage[T1]{fontenc}
\usepackage{caption}
\usepackage{subcaption}

\begin{document}
\title{Impact of network topology on the performance of Decentralized Federated Learning}
\input{parts_of_main/front_matter}
\input{parts_of_main/introduction}
\input{parts_of_main/related_work}
\input{parts_of_main/decavg}
\input{parts_of_main/body}
\input{parts_of_main/conclusion}
\input{parts_of_main/acks}


 \bibliographystyle{elsarticle-num} 
 \bibliography{cas-refs}





\end{document}

%% file: parts_of_main/front_matter.tex
\begin{frontmatter}



\author[inst1]{Luigi Palmieri\corref{cor1}} 
\ead{luigi.palmieri@iit.cnr.it}
\author[inst1]{Chiara Boldrini\fnref{label1}}
\ead{chiara.boldrini@iit.cnr.it}
\author[inst1]{Lorenzo Valerio\fnref{label1}}
 \ead{lorenzo.valerio@iit.cnr.it}
\author[inst1]{Andrea Passarella}
 \ead{andrea.passarella@iit.cnr.it}
\author[inst1]{Marco Conti}
 \ead{marco.conti@iit.cnr.it}
\affiliation[inst1]{organization={CNR-IIT},
            addressline={Via G. Moruzzi, 1}, 
            city={Pisa},
            country={Italy}}
\cortext[cor1]{Corresponding author.}
\fntext[label1]{C. Boldrini and L. Valerio contributed equally to this work.}

\begin{abstract}
Fully decentralized learning is gaining momentum for training AI models at the Internet's edge, addressing infrastructure challenges and privacy concerns. In a decentralized machine learning system, data is distributed across multiple nodes, with each node training a local model based on its respective dataset. The local models are then shared and combined to form a global model capable of making accurate predictions on new data. Our exploration focuses on how different types of network structures influence the spreading of knowledge - the process by which nodes incorporate insights gained from learning patterns in data available on other nodes across the network.
Specifically, this study investigates the intricate interplay between network structure and learning performance using three network topologies and six data distribution methods. These methods consider different vertex properties, including degree centrality, betweenness centrality, and clustering coefficient, along with whether nodes exhibit high or low values of these metrics. Our findings underscore the significance of global centrality metrics (degree, betweenness) in correlating with learning performance, while local clustering proves less predictive. We highlight the challenges in transferring knowledge from peripheral to central nodes, attributed to a dilution effect during model aggregation. Additionally, we observe that central nodes exert a pull effect, facilitating the spread of knowledge. In examining degree distribution, hubs in Barab\'asi-Albert networks positively impact learning for central nodes but exacerbate dilution when knowledge originates from peripheral nodes. Finally, we demonstrate the formidable challenge of knowledge circulation outside of segregated communities.
%
\end{abstract}



\begin{keyword}
decentralized learning \sep network topologies \sep data distribution
\end{keyword}

\end{frontmatter}

%% file: parts_of_main/introduction.tex
\section{Introduction}
\label{sec:intro}

The modern technology ecosystem is experiencing a dramatic shift characterised by the exponential growth of devices at the Edge of the network, together with the enormous expansion of data that they create. 
The most popular network paradigm for AI is currently a centralised one, that involves transferring data collected at the edge towards big data centres, where data are processed and large-scale AI models are trained. These trained models are then deployed to furnish AI-based services. Despite proving effective, this centralized approach raises several concerns related to data privacy and ownership. Even with the prospect of benefiting from centralised AI-based services, data generators at the network's edge are becoming less eager to disclose their private data to third parties. These worries are fueling a paradigm shift from centralised AI systems to decentralised ones.

%
To address these issues, solutions based on decentralised AI systems, such as Federated Learning (FL)~\cite{mcmahan_communication-efficient_2017}, have been proposed. In FL, the idea is to avoid the transfer of raw data, keeping the data locally at the edge devices. Specifically, the devices collaboratively train an AI model without sharing any raw data with each other. They only share the parameters of the AI models that are trained locally. These parameters are then aggregated to produce a better and capable \emph{global} model that is iteratively refined through successive rounds of collaboration. 

The standard definition of the FL framework assumes a starred network topology, where a central server orchestrates the entire process, coordinating the operations of the participating devices. However, the presence of a centralised controller introduces some challenges. It could potentially serve as a single point of failure, a potential bottleneck when the number of involved devices reaches millions, and an impediment to spontaneous, direct collaborations among users. 
In response to these challenges, a rising trend is emerging in favour of supporting fully decentralised variations of FL. 
Decentralised Federated Learning (DFL) represents an alternative to centralised Federated Learning. In DFL the connectivity between devices is represented by a generic graph, and the devices involved in the learning process typically collaborate only with their neighbours. 
The lack of a central controller overcomes the single point of failure problem but introduces other aspects that need to be investigated. Specifically, we claim that, in this scenario, the information locality and the network topology strongly affect the dynamics of the learning process, i.e. how fast and effective the spreading of knowledge about the class labels is.  

While previous work~\cite{sun_decentralized_2023} assumes that the network topology can be controlled by the network operator and optimized to make the learning process more efficient or scalable, we argue that complete decentralization can only be achieved by letting user devices spontaneously organize themselves. This implies that the network topology, in these settings, cannot be controlled by the operator. For example, an edge between two nodes in the graph may represent a trust relationship or willingness to cooperate. If the edges are weighted, the strength of the weights expresses the intensity of trust or cooperation. With this approach, users are free to cooperate with whomever they want, and the operator has no control over the cooperation patterns. Although this scenario poses a challenge from a learning perspective, it also fully exploits the human-centric, impromptu potential of fully decentralized learning systems.

Given these considerations, the primary research question addressed in this paper concerns the impact of network topology on the learning process within a fully decentralised learning system. 
Specifically, we examine a scenario where a group of devices collaborates to train a unified AI model in a completely decentralized environment, connected to each other through a complex network topology. Following the DFL framework, each participating device receives a set of models from its graph-connected neighbours. At first, these models undergo an initial aggregation step with the local model, usually through a weighted average, resulting in a refreshed aggregate model. Subsequently, this aggregated model is updated via a number of local training epochs performed on local data. The newly updated models are then shared with neighbouring devices, and this iterative process continues until a stopping condition is met. 

In this paper, we examine three network topologies: Erdős-Rényi, Barabási-Albert, and Stochastic Block Model. Through simulations, we analyze how their underlying network structures impact the learning process, considering various non-IID data partitioning scenarios. Specifically, we assume that a subset of nodes possesses more knowledge than others, represented by classes unavailable on other nodes. We identify this subset of nodes based on the highest or lowest values of properties such as degree centrality, betweenness centrality, and clustering coefficient. These properties capture the significance of nodes in the network and the connectivity within their local neighborhoods.
The main take-home messages from our analysis are the following.
\begin{itemize}
    \item The initial data distribution on high vs low degree nodes plays a key role in the final accuracy of a decentralized learning process.
    \item When high-degree nodes possess more knowledge, such knowledge spreads easily in the network.
    \item Vice versa, when low-degree nodes have more knowledge, knowledge spreads better when the network is less connected (at first counterintuitive, but connectivity dilutes knowledge in average-based decentralized learning).
    \item When users are grouped in segregated communities, it is very difficult for knowledge to circulate outside of the community,
    \item When considering different centrality measures, the system's performance exhibits a direct correlation with global centrality metrics. 
    In contrast, centrality measures centered on local structure are not robust indicators of system performance.
\end{itemize}

The remainder of this paper is organised as follows. We present a literature review in Section~\ref{sec:related_work}. Then, we provide the basics of Decentralized Federated Learning in Section~\ref{sec:dfl}. In Section~\ref{sec:netowork_topologies}, we discuss the network topologies that we have chosen for our analysis. The detailed settings of our experiments are discussed in Section~\ref{sec:settings}. Then, we discuss our findings for Erdős-Rényi (Section~\ref{sec:results_er}), Barabási-Albert (Section~\ref{sec:results_ba}), as well as a comparison between the two (Section~\ref{sec:er_vs_ba}). The case of segregated communities with SBM is presented in Section~\ref{sec:results_sbm}. Finally, Section~\ref{sec:conclusion} concludes the paper.

%% file: parts_of_main/related_work.tex
\section{Related work}
\label{sec:related_work}

DFL extends the typical settings of FL, e.g., data heterogeneity and non-convex optimisation, by removing the existence of the central parameter server. 
This is a relatively new topic that is gaining attention from the community since it fuses the privacy-related advantages of FL with the potentialities of decentralised and uncoordinated optimisation and learning. 
In \cite{Roy2019}, the authors define a DFL framework for a medical application where a number of hospitals collaborate to train a Neural Network model on local and private data. They make use of a decentralised Federated Learning setting where multiple medical centers can collaborate and benefit from each other without sharing data among them. In \cite{Lalitha2019} the authors propose a Bayesian-like approach where the aggregation phase is done by minimising the Kullback-Leibler divergence between the local model and the ones received from the peers. All these approaches are still considering that the nodes perform just one local update before sharing the parameters (or gradients) with the peers in the network. This aspect is relaxed in \cite{savazzi_federated_2020} and \cite{sun_decentralized_2023}. In \cite{savazzi_federated_2020}, the authors propose a federated consensus algorithm extending FedAvg from~\cite{mcmahan_communication-efficient_2017} in decentralised settings, mainly considering industrial and IoT applications. 
The authors propose a consensus-based serverless Federated Learning approach to tackle the issues that stem from massive IoT networks. The proposed FL algorithms leverage the cooperation of devices that perform data operations inside the network by iterating local computations and mutual interactions via consensus-based methods. The proposed methodology is validated on an IIoT scenario typical of a 5G network infrastructure. A slightly different approach has been proposed by authors in \cite{nguyen_--fly_2021}. They propose a variation in FL where the central server role is rotated among the participants, they refer to it as \emph{flying master}, which is dynamically selected. They demonstrate a significant reduction of runtime using their proposed flying master FL framework compared to the original FL over real 5G networks. The work in~\cite{sun_decentralized_2023} proposes a Federated Decentralised Average based on SGD in their research, where they incorporate a momentum term to counterbalance potential drift caused by multiple updates, along with a quantization scheme to reduce communication requirements. Finally, \cite{valerio2023coordination} addresses the heterogeneity and initialization problems of fully decentralized federated learning by proposing a novel learning strategy based on distance-weighted aggregation of models and knowledge distillation.

Most of the above papers assume a decentralised system made of a few nodes connected through controlled network topologies, e.g., rings and full meshes, or focuses on the algorithmic challenges of decentralised learning. 
To the best of our knowledge, \cite{palmieri2023effect} and \cite{palmieri2023exploring} were the first papers to focus on the impact of different complex network topologies on the performance of DFL. Specifically, in \cite{palmieri2023effect} we started the exploration of such relationship by observing how the data distribution, when carried out prioritizing low or high degree nodes, affects the accuracy of learning process for a set of standard network topologies such as Erdős-Rényi, Barabási-Albert, and Stochastic Block Model. In \cite{palmieri2023exploring}, we investigated the impact of network disruption during the learning process, which is an orthogonal problem to the one addressed here.
In this paper, we proceed further to investigate the relationship highlighted in \cite{palmieri2023effect}, considering additional centrality metrics for data assignment and focusing on idempotent Erdős-Rényi and Barabási-Albert graphs (this enables direct comparison between the topologies, as explained in Section~\ref{sec:network-settings}).

%% file: parts_of_main/decavg.tex
\section{Decentralised Federated Learning}
\label{sec:dfl}

Decentralised Federated Learning (DFL) is a decentralised machine learning solution that does not rely on a central coordinator.
It consists in a variation of Federated Learning (FL) that operates in a fully decentralised setting where there is no central server. In this scenario, a number $N$ of devices have to accomplish a distributed and collaborative learning task under the very same settings of federated learning\cite{verbraeken2020survey}. Basically, in Decentralised Federated Learning, there is no central entity that coordinates the nodes: the orchestrator can either not exist at all as a central entity or its role can be rotated among the participants. In a fully decentralised setting, the communication network is essentially peer-to-peer (P2P). 
The decentralised learning algorithms designed for such a P2P system are typically composed of two main blocks: one step for the local training of the model using local data and one step devoted to the exchange and aggregation of the models’ updates. These steps go on iteratively for $n$ communication rounds. 

The P2P communication network can be represented as a graph. Therefore, we model the network connecting the nodes as $\mathcal{G}(\mathcal{V},\mathcal{E})$, where $\mathcal{V}$ denotes the set of nodes and $\mathcal{E}$ the set of edges. Without loss of generality, we assume that the graph represents a social network. Exactly the same concepts can be translated to different application domains. We denote with $\omega_{ij}$ the weights on the edge between nodes $i$ and $j$ which, in the case of a social network, would represent the trust/social intimacy between the two nodes. The self-trust $\omega_{ii}$ is a pseudo-parameter with which we capture the importance placed by node $i$ on itself. We assume that only nodes sharing an edge are willing to collaborate with each other: effectively, we use the existence of a social relationship as a proxy of trust.

Each node $i \in \mathcal{V}$ is equipped with a local training dataset~$\mathcal{D}_i$ (containing tuples of features and labels $(x,y) \in \mathcal{X} \times \mathcal{Y}$) and a local model $h_i$ defined by weights $\mathbf{w}_i$, such that $h_i(\mathbf{x}; \mathbf{w}_i)$ yields the prediction of label $y$ for input $\*x$. Let us denote with $\mathcal{D} = \bigcup_i \mathcal{D}_i$ and with $\mathcal{P}$ the label distribution in $\mathcal{D}$. In general, $\mathcal{P}_i$ (i.e., the label distribution of the local dataset on node $i$) may be different from $\mathcal{P}$. This captures a realistic non-IID data distribution. 
At time 0, the model $h(\cdot; \*w_i)$ is, as usual, trained on local data, by minimizing a target loss function $\ell$ -- i.e., $\*w_i = \mathrm{argmin}_{\*w} \sum_{k = 1}^{|\mathcal{D}_i |} \ell(y_k, \*w \*x_k)$, with $(y_k, \*x_k) \in \mathcal{D}_i$.

We assume that nodes entertain a certain number of communication rounds, where they exchange and combine local models. At each communication round, a given node receives the local models from its neighbors in the social graph, and averages it with its local model. 
Specifically, at each step $t$, the local model of the given node and the local models from the node's neighbors are averaged as follows:
\begin{equation}
    \*w_i(t) \leftarrow \frac{\sum_{j \in \mathcal{N}(i)} \omega_{ij} \alpha_{ij} \*w_j(t-1)}{\sum_{j \in \mathcal{N}(i)} \omega_{ij}},
    \label{eq:weight_update}
\end{equation}
where we have denoted with $\mathcal{N}(i)$ the neighborhood of node~$i$ including itself and $\alpha_{ij}$ is equal to $\frac{|\mathcal{P}_j|}{\sum_{j \in \mathcal{N}_i} | \mathcal{P}_j|}$ (and captures the relative weight of the local dataset of node $j$ in the neighborhood of node $i$).
Once the aggregation of models is performed, the local model is trained again on the local data.

From \Cref{eq:weight_update} we can see that the aggregation step takes into consideration only the 1-hop neighboring nodes of the \emph{i-th} node (including the node itself). As a result, the topology determines the paths along which information, or ``knowledge'', travels and influences the speed and efficiency of information dissemination. The structure of the network, such as its degree distribution, clustering coefficient, and connectivity patterns, can impact the performance and effectiveness of decentralised federated algorithms.
Thus, it is reasonable to expect the network topology to play a crucial role in DFL. This is the rationale for analyzing the impact of network topology in such settings.

%% file: parts_of_main/body.tex
\section{Network Topologies}
\label{sec:netowork_topologies}

In this paper, we analyze three different network topologies, i.e., Erdős-Rényi (ER), Barabási-Albert (BA), and  Stochastic Block Model (SBM), to investigate how their properties impact on the diffusion of  knowledge during a fully decentralized learning process. We consider non-directed graphs, assuming that node communication is always bidirectional. 

The ER model is a model for generating random graphs with a homogeneous structure, where nodes are connected to each other with a fixed probability. 
ER is defined by two parameters: $N$, the number of nodes in the network, and $p$, the probability of an edge existing between any two nodes in the network (regardless of their degree). The ER model shows a phase transition when the fixed probability $p$ approaches the critical value $p^*=\ln(N)/N $~\cite{erdHos1960evolution}. Specifically, the value $p^*$ is a sharp threshold for the connectedness of the network: for values of $p$ above $p^*$ the network goes from presenting isolated nodes to almost surely be made of a unique connected component such that all nodes are reachable within a finite number of hops.

The BA is an algorithm for generating random scale-free networks, i.e., networks with a power-law (or scale-free) degree distribution, using a preferential attachment mechanism~\cite{albert2002statistical}. In the BA model, nodes are connected preferentially based on their degree. Specifically, the probability of an edge forming between two nodes is proportional to the nodes' degree, which leads to the emergence of a scale-free degree distribution. Since the degree distribution follows a power law, few nodes have a very high degree while most nodes have a low degree. This can result in a structure with few well-connected hubs, which are known to facilitate information flow across the network. A BA network is defined by two parameters: $N$, the number of nodes in the network, and $m$, the number of edges added to the network for each new node (hence, the minimum degree of nodes). 

The SBM is a probabilistic model for networks that exhibit a modular structure, i.e., the SBM generates a network with a clear community structure where nodes are grouped together based on their connectivity patterns~\cite{lee2019review}.
Nodes belonging to the same group are more closely connected to each other than to nodes in another group.
Formally, the SBM is defined by the following parameters: $N$, the number of nodes in the network;  $B$, the number of communities (called blocks); ${n_1, n_2, ..., n_B}$, the sizes of the blocks where $n_i$ is the number of nodes in block $i$; $p_{ij}$, the probability of an edge existing between a node in block $i$ and a node in block $j$ (with $p_{ii}$ the probability of links inside the block).

These three models capture important properties of complex networks. ER networks, which are random and well-mixed, provide insights into how information propagates in networks without a pronounced structure, with homogeneity in terms of degree and low clustering coefficient. BA graphs are characterised by a very skewed degree distribution with few high-degree nodes and many low-degree nodes. The presence of nodes with high degree can enhance rapid dissemination, contributing to efficient diffusion, but may overshadow the contribution of low degree nodes. Finally, SBM introduces the concept of community structure. SBM feature a well-defined community structure that allows us to investigate how knowledge spreads within and between communities.

\section{Experimental settings}
\label{sec:settings}


\subsection{Network settings}
\label{sec:network-settings}

In this paper, we consider unweighted graphs with 100 nodes, where edges are generated as follows. 

\begin{figure}[p!]
    \begin{minipage}{\textwidth}
    \centering
    \includegraphics[width=\textwidth]{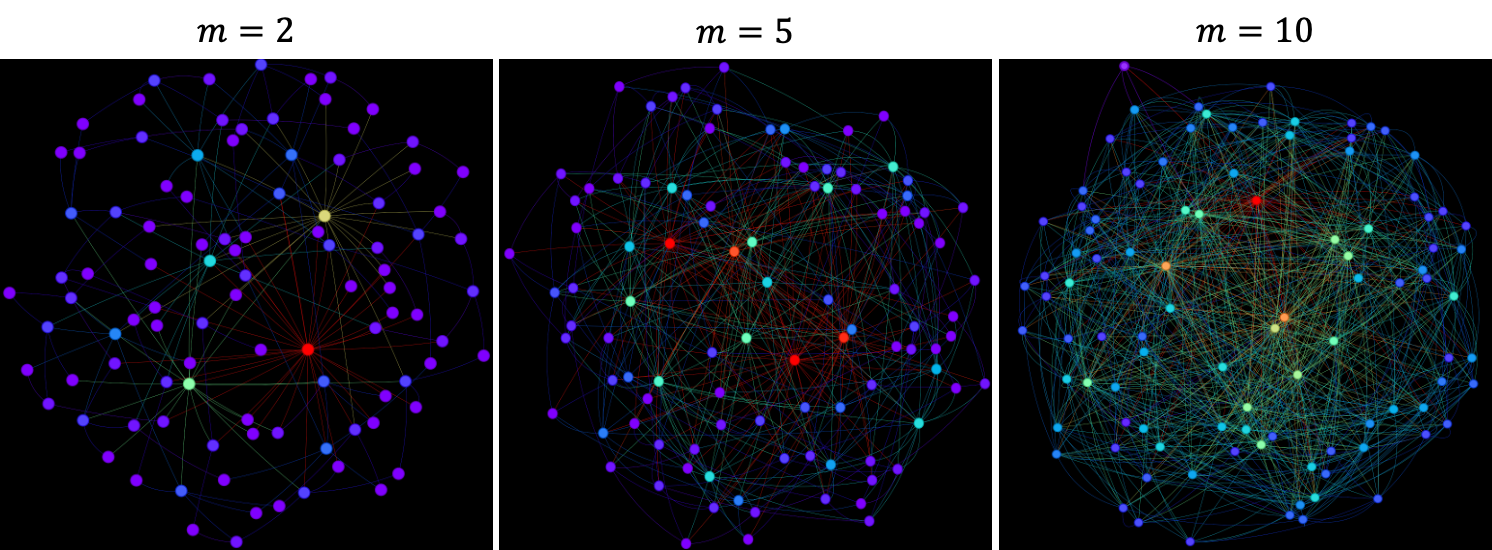}
    \caption{The three BA networks considered in our experiments, with increasing value of the preferential attachment parameter $m$. Vertices are colored based on their degree, going from purple to red with increasing degree values.}
    \label{fig:ba_graphs}
    \end{minipage}
%
    \begin{minipage}{\textwidth}
    \centering
    \includegraphics[width=\textwidth]{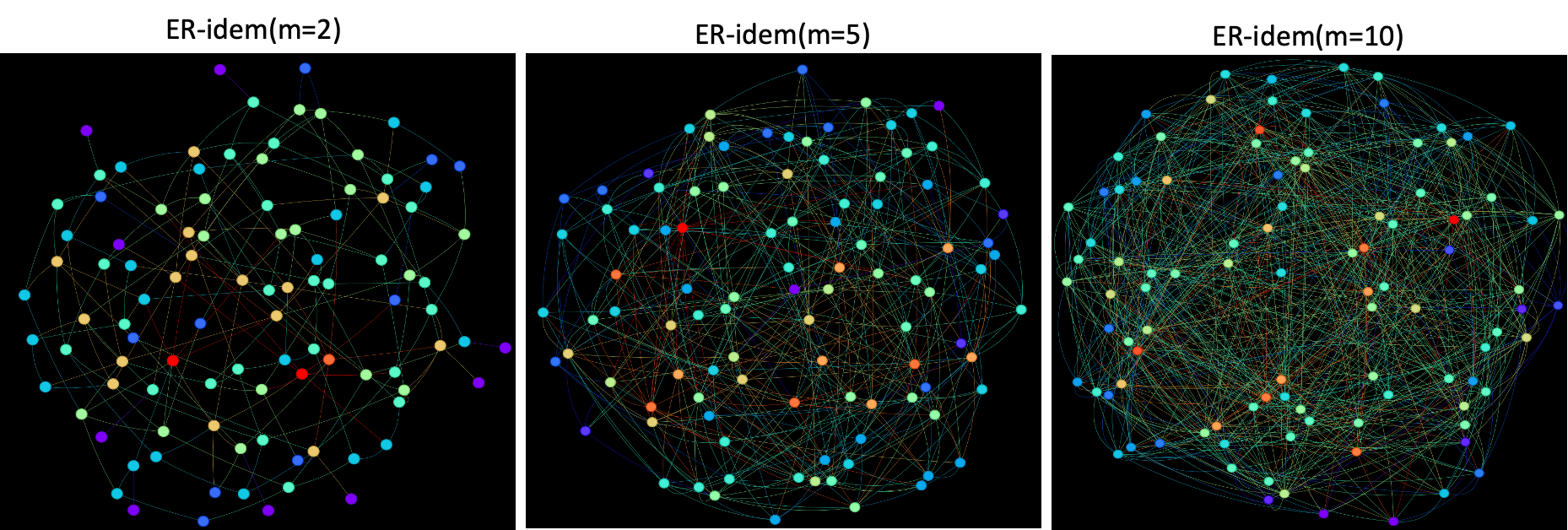}
    \caption{The three different realizations of the Erdős-Rényi graph, idempotent to the corresponding BA graph with increasing value of the preferential attachment parameter $m$. Vertices are colored based on their degree, going from purple to red with increasing degree values.}
    \label{fig:er_original}
    \end{minipage}
%
    \begin{minipage}{\textwidth}
    \centering
    \includegraphics[width=.7\textwidth,height = 5cm]{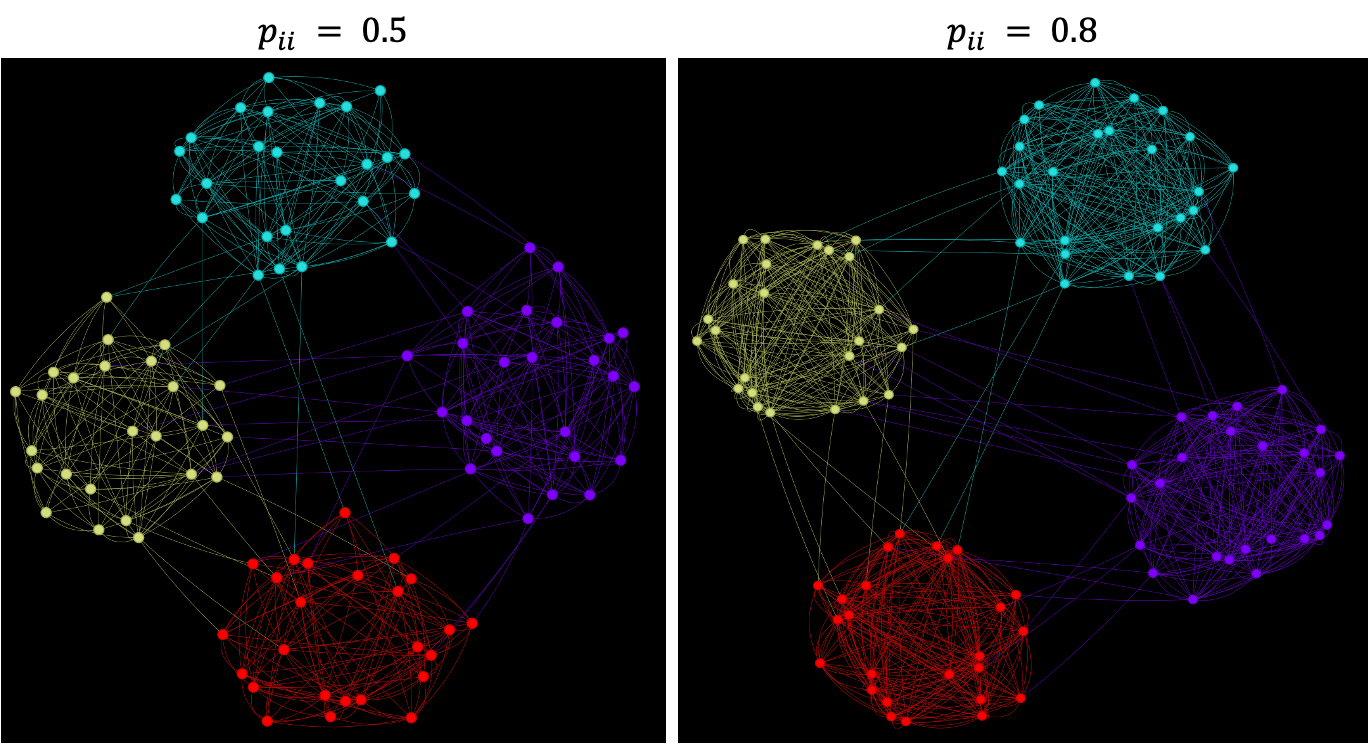}
    \caption{The two SBM graphs. Colouring is based on the data distribution as described in the body of the paper.}
    \label{fig:sbm_all}
    \end{minipage}
\end{figure}

\noindent\textbf{Barabási-Albert.} Three different cases regarding the parameter of preferential attachment are chosen: $m = 2,5,10$, leading to networks with increasingly higher node degrees, as illustrated in~\Cref{fig:ba_graphs}.

\noindent\textbf{Erdős-Rényi.} In order to draw direct comparisons against BA networks, for each BA graph we generate the corresponding idempotent ER network\footnote{Note that this implies that the ER graphs considered in this paper are different from those studied in~\cite{palmieri2023effect}.}, i.e., an ER network with the same number of nodes and edges with respect to its corresponding BA network. To this aim, for each BA network, we randomly reshuffled the links between the nodes keeping the density of the original BA networks. The reshuffling was done by employing the corresponding built-in function of the \texttt{graph-tool} python library~\cite{peixoto_graph-tool_2014}. We denote with ER-idem($m$) the ER idempotent to the BA graph with parameter $m$. In order to understand where the generated ER graphs are positioned with respect to the critical threshold (see discussion in Section~\ref{sec:netowork_topologies}), the equivalent $p$ values of the ER-idem($m$) are shown in~\Cref{tab:equiv_er_p}. For the $m=2$ case, the corresponding ER graph will have a $p$ value close to the critical one for the connectedness of the network, which is $p^* = 0.046$ (obtained from $\frac{\ln{100}}{100}$ \cite{erdHos1960evolution}). For the other networks, the values are well above the critical value for connectedness. An increase in the values of $p$ when the number of nodes stays fixed results in a higher density of edges, as shown  in~\Cref{tab:summary-networks}, which reports the main network indices for the considered graphs. 

The clustering coefficient is higher for BA while the average shortest path length and the diameter tend to be higher for ER. Recall that a higher clustering coefficient implies that neighbourhoods are better connected internally, while a higher average shortest path length implies that it takes more hops, on average, to reach a generic node in the graph (a similar consideration holds for the diameter). 

\begin{table}[t!]
\centering
\begin{tabular}{@{}lll@{}}
\toprule
\textbf{ER-idem($m=2$)} & \textbf{ER-idem($m=5$)} & \textbf{ER-idem($m=10$)} \\ \midrule
$p=0.0396$     & $p=0.0960$    & $p=0.182$      \\ \bottomrule
\end{tabular}
\caption{Equivalent values of $p$ for the ER networks idempotent to the corresponding BA networks.}
\label{tab:equiv_er_p}
\end{table}



\noindent\textbf{Stochastic Block Model.} Nodes are grouped into 4 communities (denoted as $C_1,C_2,C_3,C_4$) of equal size (25 nodes each). The probability of extrinsic connections $(p_{ij}, \; j \neq i)$ is set to~0.01, whereas the probability of intrinsic connections $(p_{ii})$ is set to 0.8 in one case and 0.5 in the second case. The resulting graphs are plotted in~\Cref{fig:sbm_all}. The reason for such values of intrinsic connectivity are to explore the impact of a more loose or more tightly connected community structure. The level of connectivity obtained by the $p_{ii} =0.5$ case represent a balanced or moderate degree of interconnectedness within the communities of the network. Whereas the intrinsic connectivity of 0.8 indicates a scenario where the connections within the blocks are denser (closer to 1).

\begin{table}[t!]
\centering
\scriptsize
\begin{tabular}{@{}lrrrrrrrr@{}}
\toprule
\multicolumn{1}{l}{} & \multicolumn{3}{c}{\textbf{Barabási-Albert}}   & \multicolumn{3}{c}{\textbf{Erdős-Rényi}}    & \multicolumn{2}{c}{\textbf{SBM}} \\  
\multicolumn{1}{l}{} & \multicolumn{3}{c}{$m$}   & \multicolumn{3}{c}{$\textrm{idem}(m)$}    & \multicolumn{2}{c}{$p_{ii}$} \\
\cmidrule(lr){2-4}  \cmidrule(lr){5-7}  \cmidrule(lr){8-9} 
  & $2$ & $5$ & $10$ & $2$ & $5$ & $10$ & $0.5$ & $0.8$ \\
\midrule
\textbf{Number of nodes} & 100 & 100& 100& 100& 100& 100& 100& 100 \\
\textbf{Number of edges} & 196 & 475 & 900 & 196 & 475 & 900 & 624 & 1003 \\
\textbf{Avg degree} & 3.92 & 9.50 & 18.00 & 3.92 & 9.50 & 18.00 & 12.48 & 20.06 \\
\textbf{Density} & 0.0396 & 0.0960 & 0.182 & 0.0396 & 0.0960 & 0.182 & 0.126 & 0.203 \\
\textbf{Clustering coef.} & 0.164 & 0.213 & 0.296 & 0.0394 & 0.0900 & 0.194 & 0.424 & 0.747 \\
\textbf{Avg shortest path l.} & 2.85 & 2.17 & 1.85 & 3.49 & 2.28 & 1.85 & 2.56 & 2.30 \\
\textbf{Diameter} & 5 & 3 & 3 & 7 & 4 & 3 & 5 & 4 \\
\bottomrule
\end{tabular}
\caption{Summary statistics for the considered networks.}
\label{tab:summary-networks}
\end{table}

\begin{figure}[t!]
    \centering
    \includegraphics[width=\textwidth]{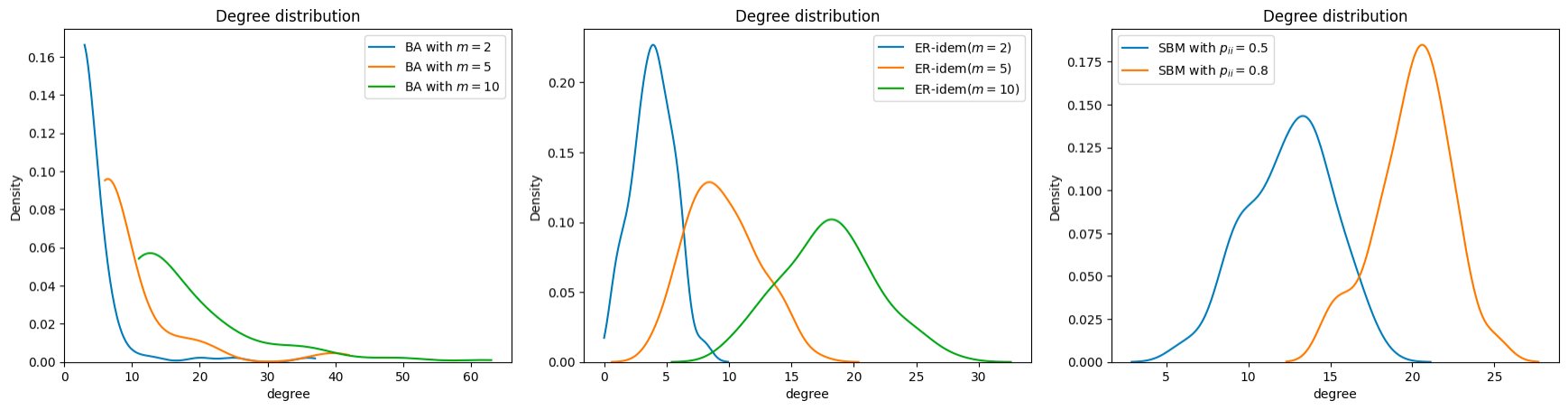}
    \caption{Degree distributions for the analyzed networks. From left to right: Barabási-Albert, Erdős-Rényi and Stochastic Block model.}
    \label{fig:degree_dist}
\end{figure}

\subsection{Data distribution}
\label{sec:data_dist}
For our experiments, we choose the widely used MNIST image dataset~\cite{lecun1998mnist}. We selected MNIST because it is a well-understood dataset that allows us to focus our analysis on the dynamics induced by the network topology.  This dataset contains a set of handwritten digits, thus data are divided into 10 classes. The goal of the analysis is to characterise the effect of the network topology in the knowledge spreading process, i.e., the ability of nodes to learn data patterns they have \emph{not} seen locally, but that other nodes in the network have seen. Therefore, we  split the MNIST classes across nodes as follows. Note that, on the assigned classes, each node gets the same amount of images.

For ER and BA networks, we divide the 10 MNIST classes into two groups: the first group (G1) is composed of classes 0, 1, 2, 3, 4, and the second group (G2) of classes 5, 6, 7, 8, 9. All nodes receive an equal share (selected randomly) of data from G1. Data from G2 are allocated only to a subset of nodes, selected as follows. for each type of network measure considered, we examine two cases where data in G2 are assigned to the 10\% of nodes with the highest values and the 10\% with the lowest values, respectively.  The rationale is thus to allocate ``full knowledge" (i.e., a complete subset of all classes) either to high-centrality or low-centrality nodes, and study the effect of the network topology in both cases. In the following, these configurations are referred to as ``highest-focus" and ``lowest-focus", respectively. 
The measures taken into account are:
\begin{itemize}
    \item Degree centrality
    \item Betweenness centrality
    \item Clustering coefficient 
\end{itemize}
For the degree centrality measure, we will refer to the highest-focus and lowest-focus cases as hub-focus and edge-focus to easily distinguish the degree results with the other centrality measures.
Specifically, starting from the node(s) with the highest (lowest) centrality, we pick nodes until we reach 10\% of the network. In case adding all nodes at a given centrality measure value results in more than 10\% of the network, we randomly pick, among nodes with that centrality measure value, a subset that allows us to fill the 10\% subset. 

The \emph{degree centrality} is calculated by taking the total number of edges connected to a node (i.e., its degree). Being directly connected with many other nodes, nodes with high degree centrality are generally expected to be more influential in the network, as they can reach many nodes at once leveraging their direct connections. When considering the degree centrality, low-centrality nodes are edge nodes (i.e., nodes at the periphery of the graph), hence we may use the terms lowest-focused and edge-focused interchangeably. Similarly for highest-focused and hub-focused.

The \emph{betweenness centrality} measures the importance of a node in a network by quantifying its role as a connector or bridge between other nodes~\cite{freeman1977set}. It identifies nodes that lie on a high number of shortest paths, which are the most efficient routes for information to travel between different parts of the network. These nodes play a critical role in facilitating communication and information flow, as they act as intermediaries between different communities or groups of nodes. Nodes with high betweenness centrality are often referred to as \textit{bridges}. In contrast, nodes with low betweenness centrality lie on relatively few shortest paths and are less influential in connecting different parts of the network. They are often referred to as
\textit{isolates}.

The \textit{clustering coefficient} of a node measures the proportion of its neighbours that are also connected to each other~\cite{barabasi2013network}. Nodes with higher clustering coefficient centrality have neighbours that are well-connected to each other, indicating the presence of cohesive groups or clusters in the network. The higher the clustering coefficient, the more homogeneous a group of nodes is in terms of connections, the more balanced information spread is expected to be within the cluster.

The rationale behind the choice of these centrality measures is their ability to capture distinct aspects of network dynamics and structure. Degree centrality highlights the importance of well-connected nodes, clustering coefficient emphasizes local connectivity patterns, and betweenness centrality identifies nodes critical for global information flow. 

For SBM networks, instead, we divided the dataset classes into subsets based on the communities the nodes belong to, without overlap. Therefore, since we study SBM topologies with 4 communities, each community gets two classes: community 1 sees classes 0 and 1; community 2 sees classes 2 and 3; community 3 sees classes 4 and 5; community 4 sees classes 6 and 7. This data distribution is designed to challenge the knowledge-spreading process since maximum learning accuracy can only be achieved if information from all the external communities is brought into the local one. 

\subsection{Learning task and evaluation metric}
\label{sec:eval_metrics}
For the learning task, we consider a simple classifier as model to be trained and we focus on two performance figures. On the one hand, we consider the accuracy over time at each node, to assess the effectiveness and speed of knowledge diffusion across the network. 
On the other hand, for SBM networks, we also investigated the average confusion matrix across nodes of the same community. Specifically, for each node we compute the confusion matrix for the MNIST classes, and then take the average across all nodes in the same community.

We implemented the \textsc{DecAvg} scheme within the custom SAISim simulator, available on Zenodo\footnote{\url{https://doi.org/10.5281/zenodo.5780042}}. SAISim is developed in Python and leverages state-of-the-art libraries such as PyTorch and NetworkX for deep learning and complex networks, respectively. On top of that, SAISim implements the primitives for supporting fully decentralized learning. The local models of nodes are Multilayer Perceptrons with three layers (sizes 512, 256, 128) and ReLu activation functions. SGD is used for the optimization, with learning rate~$0.01$ and momentum~$0.5$. The number of local training epochs between each round of updates' exchange is set to $100$.

\section{Results}
\label{sec:results}

The rest of this section is organised as follows: in \Cref{sec:results_er} and \Cref{sec:results_ba}, we present individually the results obtained for the ER and BA networks. Then in~\Cref{sec:er_vs_ba} we discuss how the two topopologies compare against each other. Finally, in~\Cref{sec:results_sbm} we present and discuss the results obtained for the SBM networks.

\subsection{Decentralised learning over Erdős-Rényi graphs}
\label{sec:results_er}

\begin{figure}[t!]
    \begin{minipage}{\textwidth}
    \centering
    \includegraphics[width = \textwidth,keepaspectratio]{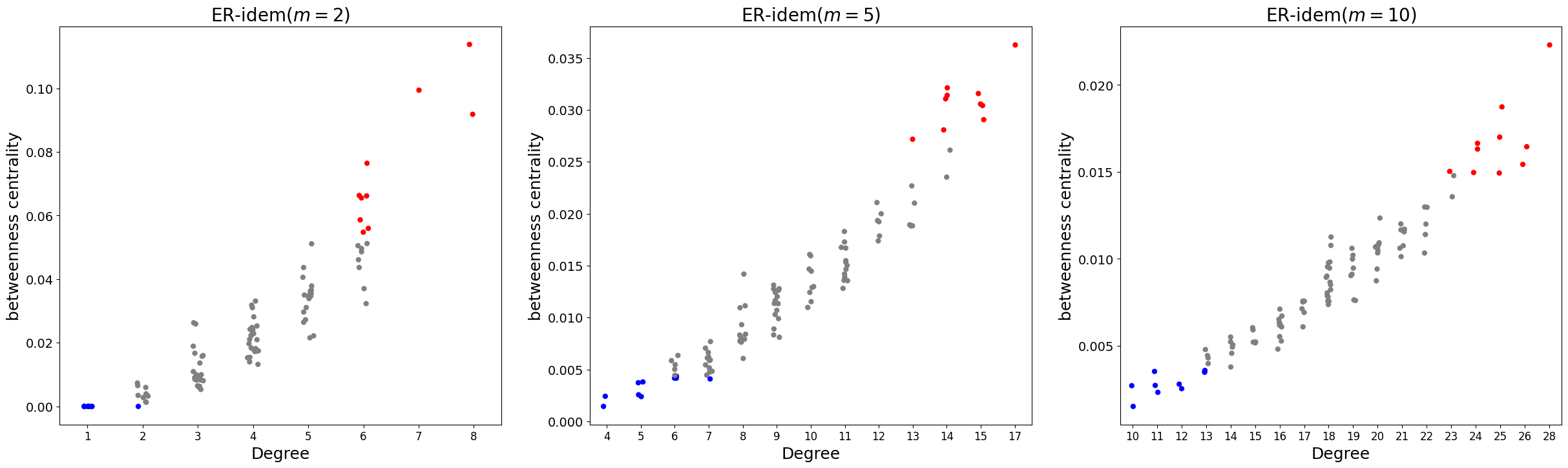}
    \caption{Scatterplots of the nodes' betweenness centrality measures vs the degree in our ER graphs. Increasing value of connectedness going from left to right. Red dots are the nodes chosen for the highest-focused case, the blue dots are the nodes chosen for the lowest-focused case. The scatterplots show that the betweenness centrality is proportional to the degree.}
    \label{fig:er_betweenness_scatterplot}
    \end{minipage}
%
    \begin{minipage}{\textwidth}
    \centering
    \includegraphics[width=\textwidth,keepaspectratio]{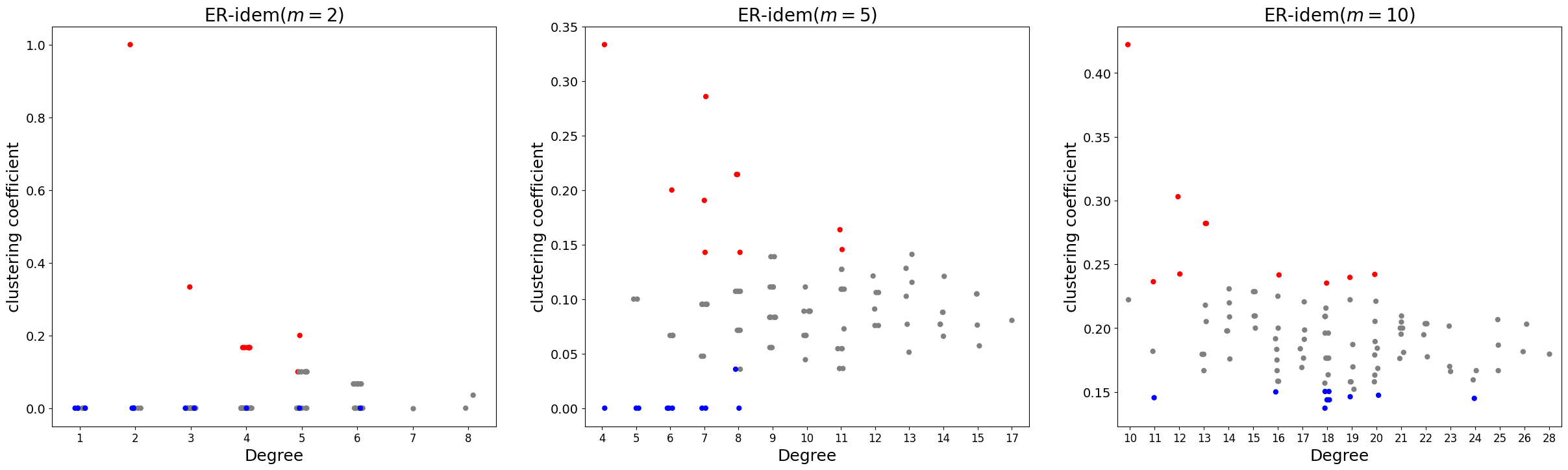}
    \caption{Scatterplots of the nodes' clustering coefficient vs the degree in our ER graphs. Increasing values of connectedness going from left to right. Red dots are the nodes chosen for the highest-focused case, the blue dots are the nodes chosen for the lowest-focused case. There is no strong correlation between the two metrics.}
    \label{fig:er_clust_scatterplot}
    \end{minipage}
\end{figure}

As explained in~\Cref{sec:settings}, for the ER model we consider three settings that are idempotent to the chosen Barabási-Albert graphs, discussed later on in~\Cref{sec:results_ba}. This allows us to draw direct comparisons between the main characteristics of a decentralized learning process on equivalent topologies (Section~\ref{sec:er_vs_ba}). 
Let us begin our analysis by considering the relationships between the centrality measures w.r.t. the considered network topology, as shown in the scatterplots of Figures~\ref{fig:er_betweenness_scatterplot}-\ref{fig:er_clust_scatterplot}. 
The scatterplots in~\Cref{fig:er_betweenness_scatterplot} show the relation between degree centrality and betweenness centrality for the three distinct configurations of the Erdős-Rényi (ER) graph. In these scatterplots, the highest-focus and lowest-focus nodes selected for the betweenness centrality metric are highlighted in red and blue, respectively.
The scatterplots show a relationship of direct proportionality between betweenness centrality and degree. Nodes with higher degrees tend to exhibit higher betweenness centrality, indicating that well-connected nodes also play crucial roles as bridges or intermediaries in the network. 
Hence, we expect a similar behaviour of the system when data is distributed either according to the betweenness centrality or their degree. 
\Cref{fig:er_clust_scatterplot} shows the scatterplots of the clustering coefficient vs the degree centrality: as it is common in such graphs, the higher the clustering coefficient, the lower the degree. 
The intuitive understanding behind nodes with high clustering coefficients having smaller degrees lies in the fact that high-degree nodes, connected to numerous other nodes, are less likely to have tightly knit connections among their neighbours, causing them to cluster in the lower degree range of the plot.

\subsubsection{ER: Degree centrality used for G2 data assignment}
\label{sec:results_er_degree}

\begin{figure}[p!]
    \begin{minipage}{\textwidth}
    \centering
    \includegraphics[width=\textwidth]{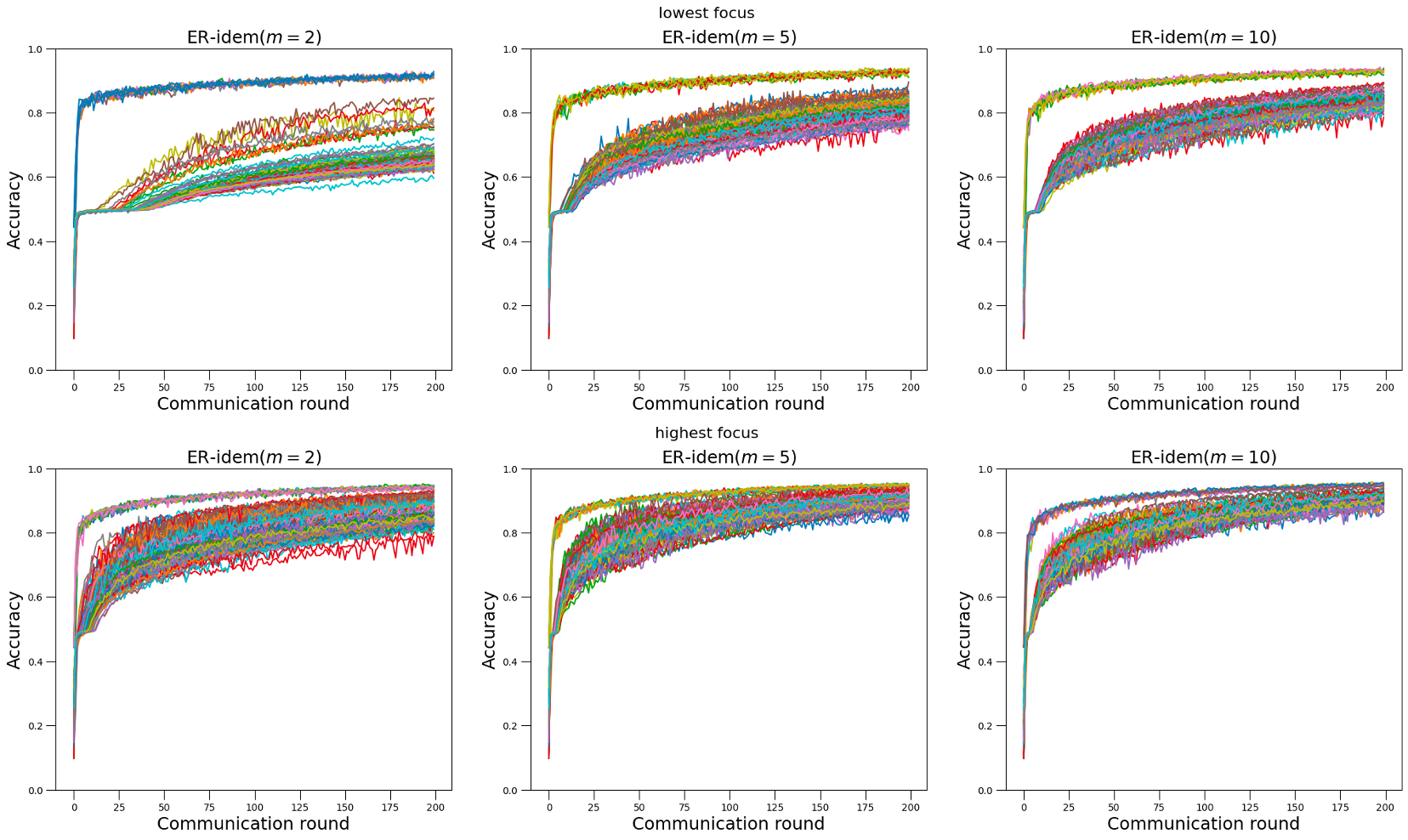}
    \caption{Accuracy over time in ER networks (all nodes, G2 data assigned based on degree centrality). From left to right: increasing values of connectedness; from top to bottom: lowest-focused and highest-focused scenario.}
    \label{fig:er_deg_all}
    \end{minipage}
%
    \begin{minipage}{\textwidth}
    \centering
    \includegraphics[width=\textwidth]{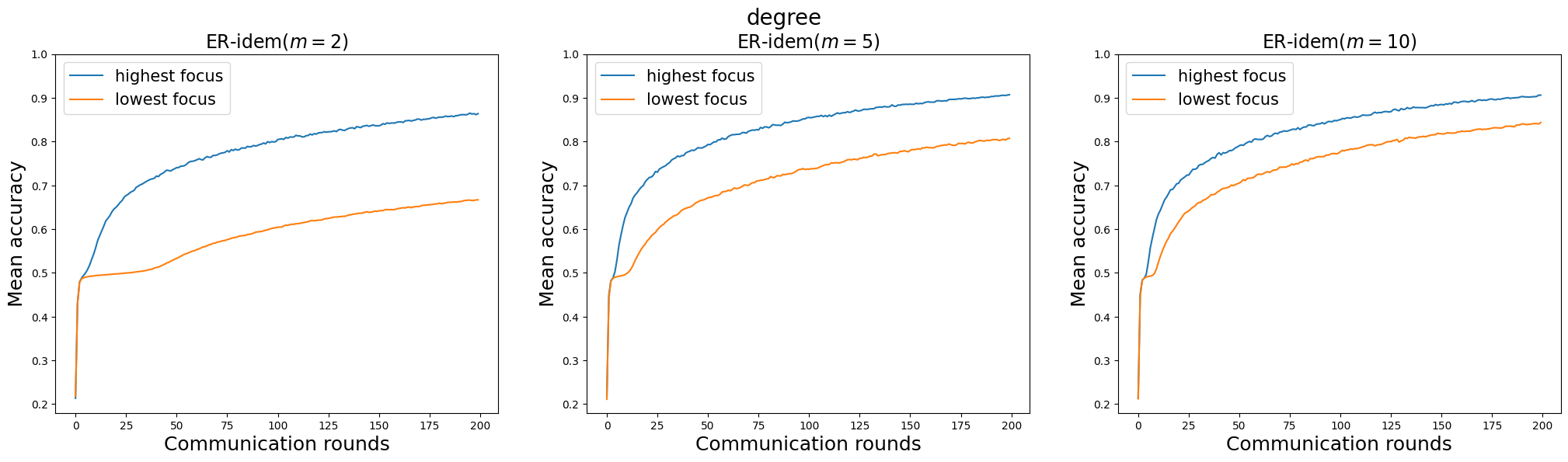}
    \caption{Mean accuracy over time in ER networks (average across G1 nodes only, G2 data assigned based on degree centrality). Highest-focused (blue) and lowest-focused (orange) cases.}
    \label{fig:er_deg_high_vs_low}
    \end{minipage}
%
    \begin{minipage}{\textwidth}
    \centering
    \includegraphics[width=0.8\textwidth,keepaspectratio]{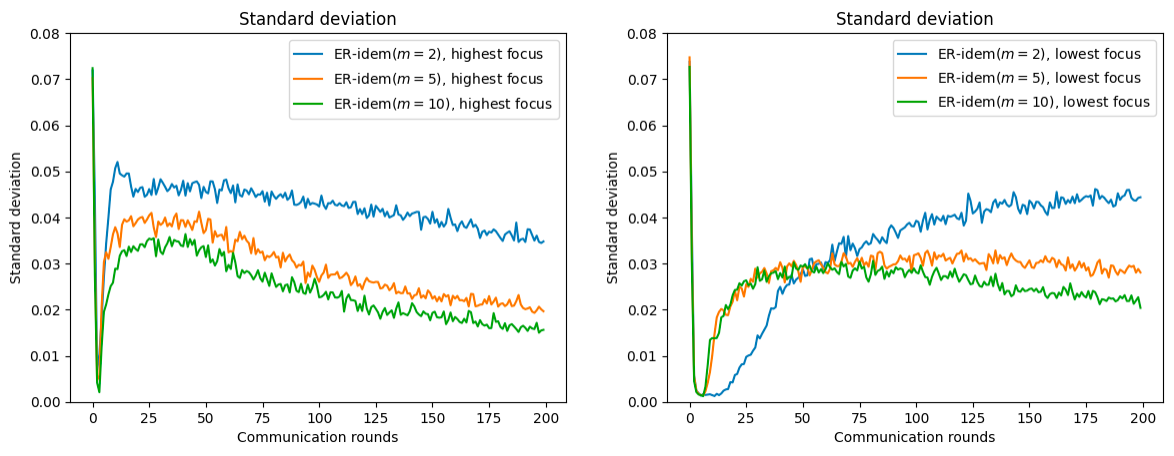}
    \caption{Standard deviation of accuracy over time in ER networks (std dev. across G1 nodes only, G2 data assigned based on degree centrality). Left to right: highest-focus and lowest-focus cases.}
    \label{fig:er_deg_std_all}
    \end{minipage}
\end{figure}

We start our analysis with the case of data in the G2 subset being assigned based on the degree centrality. For clarity, we denote nodes exclusively assigned G1 data as \emph{G1 nodes} and, conversely, refer to nodes assigned both G1 and G2 data as \emph{G2 nodes}.
In~\Cref{fig:er_deg_all} we show the evolution of the accuracy per node over time (each curve corresponds to one node). The top-row plots refer to the edge-focused case, where the digits in class group G2 are assigned to edge nodes, while the bottom-row plots correspond to the hub-focused scenario, where the high-degree nodes get the G2 class data. As expected, the nodes that are assigned both G1 and G2 data (the group of curves with higher accuracy in the figure) show an excellent learning performance. It is more interesting to focus our attention on the other nodes, the ones assigned only data in G1 that can only learn the classes in G2 thanks to the knowledge (models) they receive from other nodes. In the bottom row of~\Cref{fig:er_deg_all} we can observe the accuracy over time when G2 data are assigned to the nodes with the highest degree. Since a high degree implies better connectivity, we expected it to be easier for edge nodes to be reached by the ``good'' models of the central nodes, and this to be increasingly true as the average degree of the network increases (from left to right). This is confirmed in the figure, where we see that some edge nodes are struggling on ER-idem($m=2$) but progressively catch up when $m$ increases. \ul{In other words, when the highest degree nodes enjoy higher accuracy, they are able to drag all the other nodes closer to their performance efficiently.}

Conversely, when G2 data are instead assigned to nodes with the smallest degree, we expect the decentralized learning process to be less effective. The intuition is that nodes with G2 data, in this case, are poorly connected, hence the knowledge they bring has a hard time percolating through the network. The top row of~\Cref{fig:er_deg_all} confirms this. Especially when the average degree is smallest, corresponding to ER-idem($m=2$), we see nodes that, after 200 communication rounds, have barely increased their accuracy above the baseline $0.5$ (corresponding to the accuracy on the G1 class group that is available locally to all nodes). When comparing to the previous case, then, we conclude that \ul{when G2 nodes are less central (according to the degree centrality) they have difficulties in dragging G1 nodes towards better accuracy by contributing their ``good'' models.} \ul{This effect arises from the combination of the average-based aggregation strategy in \textsc{DecAvg} along with the topological properties of the graphs.} 

In \Cref{fig:er_deg_high_vs_low} we aggregate the accuracy among all G1 nodes in the same experiment and we show the evolution over time of the average of the accuracy. Coherently with the reasoning above, the average accuracy in the edge-focus case is much lower than in the hub-focus case. The gap decreases as we increase the overall connectivity of the network, i.e., as $m$ increases. The poor average performance of the edge-focus case is due to high-degree nodes hindering the spreading of knowledge from edge nodes. This effect is due to the model averaging mechanism of decentralized learning, as we explain in more detail in Section~\ref{sec:results_ba}.
In \Cref{fig:er_deg_std_all}, we present the standard deviation of the accuracy for all G1 nodes in the ER networks. As we can see, the hub-focused case (left) shows separated curves having a clear gap between ER-idem($m=2$) and the others. This higher variability can be attributed to the effect of longer paths of ER-idem($m=2$). Since the equivalent $p$ value of ER-idem($m=2$) is around the critical value for the connectedness, the network has longer paths, impeding the synchronization of local models across nodes. In this scenario, nodes are situated farther apart compared to the other more connected networks (values of $p$ well above the critical value). 
In the edge-focus case, we observe two separate behaviours. When the connectivity is high enough, i.e., for  ER-idem($m=5$) and ER-idem($m=10$), the trend of the standard deviations is more similar to the hub-focus case, with a decrease over time although at a slower pace. For the case ER-idem($m=2$), the trend of the variability is increasing in the observed time window, proving that the connectivity is insufficient to let the all nodes synchronize and collaborate constructively within a reasonable timeframe.

\subsubsection{ER: Betweenness centrality used for G2 data assignment}
\label{sec:results_er_betweenness}

\begin{figure}[p!]
    \begin{minipage}{\textwidth}
    \centering
        \begin{minipage}{\textwidth}
        \includegraphics[width=\textwidth,keepaspectratio]{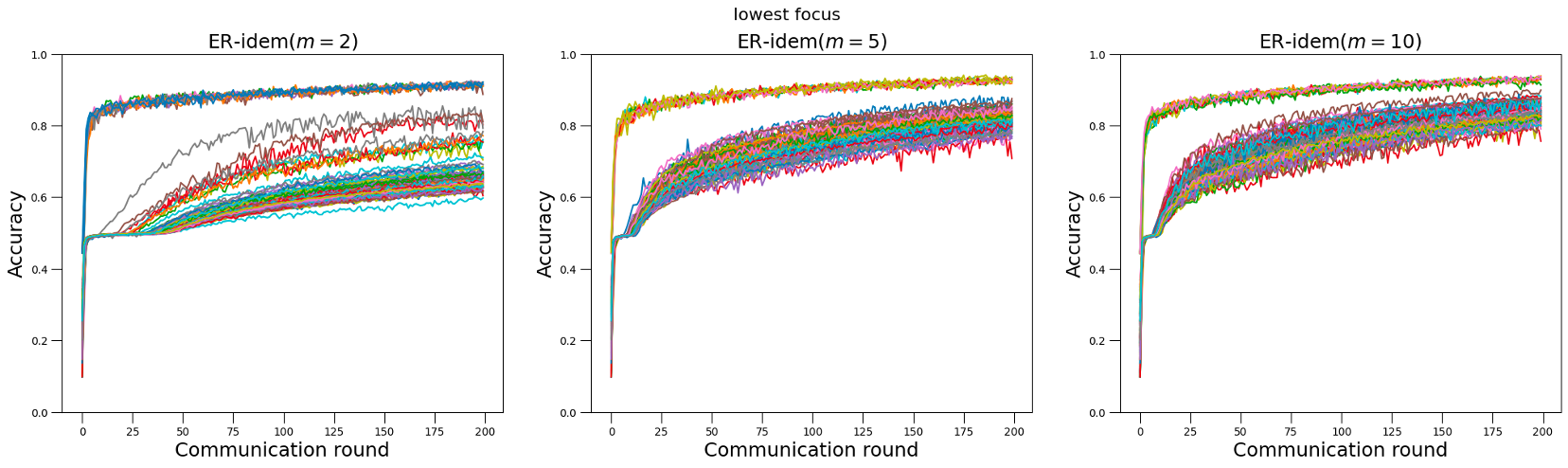}
        \end{minipage}
        \begin{minipage}{\textwidth}
        \includegraphics[width=\textwidth,keepaspectratio]{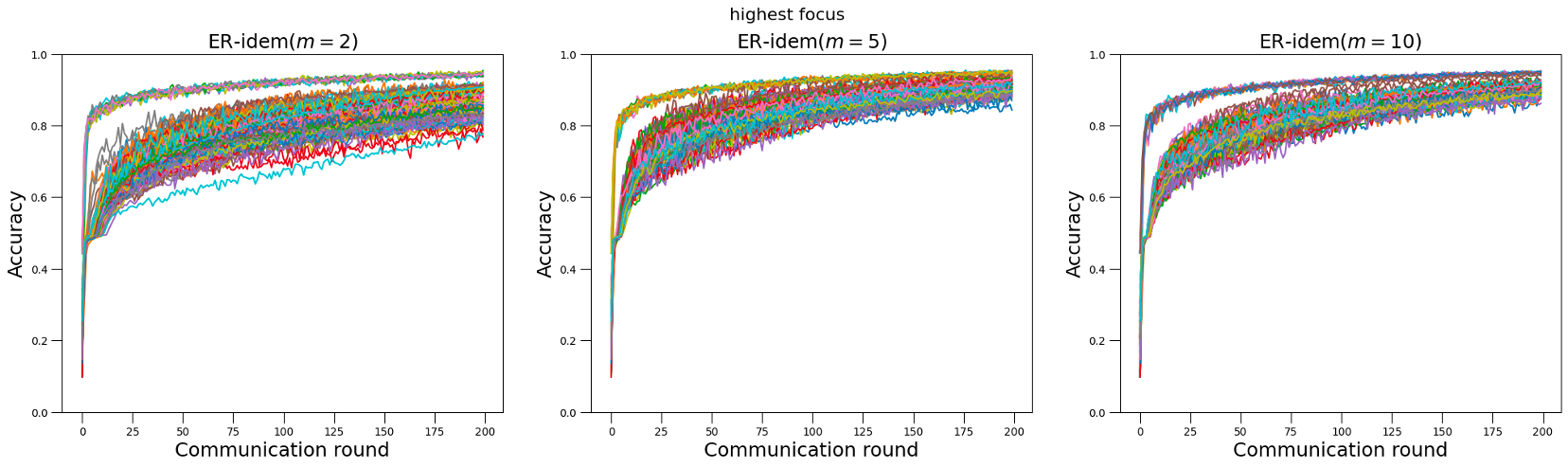}
        \end{minipage}
    \caption{Accuracy over time in ER networks (all nodes, G2 data assigned based on betweenness centrality). From left to right: increasing values of connectedness; from top to bottom: lowest-focused and highest-focused scenario.}
    \label{fig:er_bet_all}
    \end{minipage}
%
    \begin{minipage}{\textwidth}
    \centering
    \includegraphics[width=\textwidth,keepaspectratio]{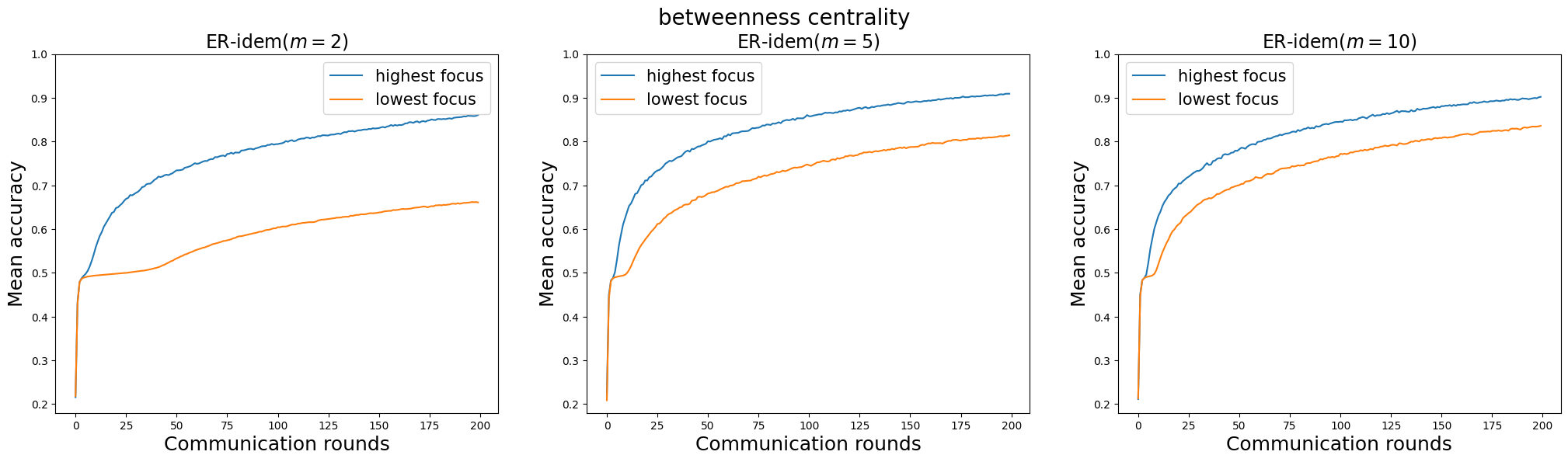}
    \caption{Mean accuracy over time in ER networks (average across G1 nodes only, G2 data assigned based on betweenness centrality). Highest-focused (blue) and lowest-focused (orange) cases.}
    \label{fig:er_bet_high_vs_low}
    \end{minipage}
%
    \begin{minipage}{\textwidth}
    \centering
    \includegraphics[width=0.8\textwidth,keepaspectratio]{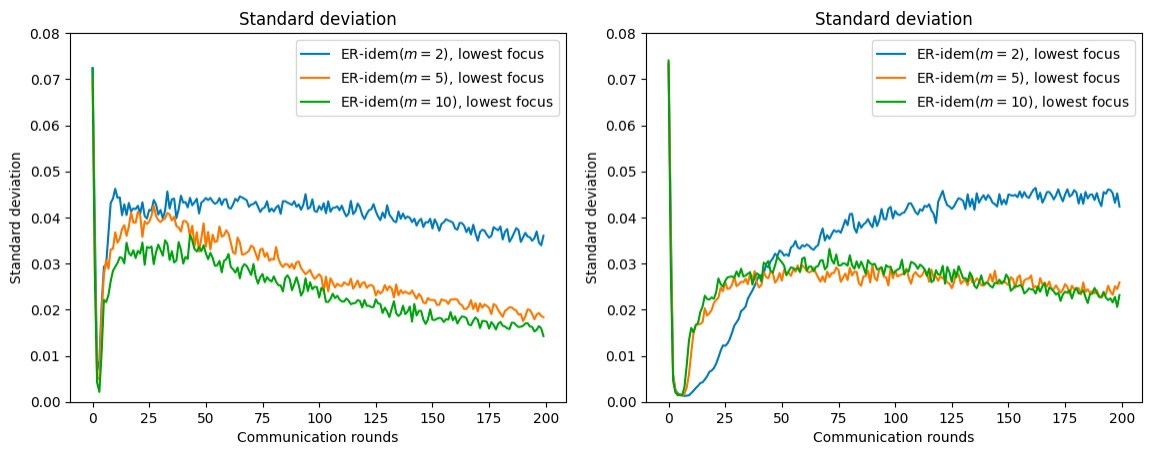}
    \caption{Standard deviation of accuracy over time in ER networks (std dev. across G1 nodes only, G2 data assigned based on betweenness centrality). Left to right: highest-focus and lowest-focus cases.}
    \label{fig:er_bet_std_all}
    \end{minipage}
\end{figure}

In this section, we investigate the accuracy performance of decentralized learning when G2 data are assigned based on the betweenness centrality. 
In~\Cref{fig:er_bet_all} we show the evolution of the accuracy per node over time. Due to the high correlation between degree centrality and betweenness centrality in our ER graphs (\Cref{fig:er_betweenness_scatterplot}), \ul{a similar behaviour as in the degree-based case emerges}. 
%
%
Consistently with the results obtained in the previous section, the highest-focused case has a better performance than the lowest-focused case. 

In \Cref{fig:er_bet_high_vs_low}, we show the average accuracy of all the nodes with assigned G1 images for both the highest and lowest focus cases. As expected, the highest focus case shows higher accuracy levels than the lowest focus counterpart and an overall small difference in performance. 
Looking at \Cref{fig:er_bet_std_all}, we obtain results similar to the degree-based analysis. Also in this case, the connectedness of the graph influence the gap between the curves for the ER-idem($m=2$) case and the other cases. As we increase the mean connectivity of the graph not only the gap decreases but so does the standard deviation, since the augmented connectivity helps the other nodes to spread their information to the other nodes.


\subsubsection{ER: Clustering coefficient used for G2 data assignment}
\label{sec:results_er_clustering}

\begin{figure}[p!]
    \begin{minipage}{\textwidth}
    \centering
    \includegraphics[width=\textwidth]{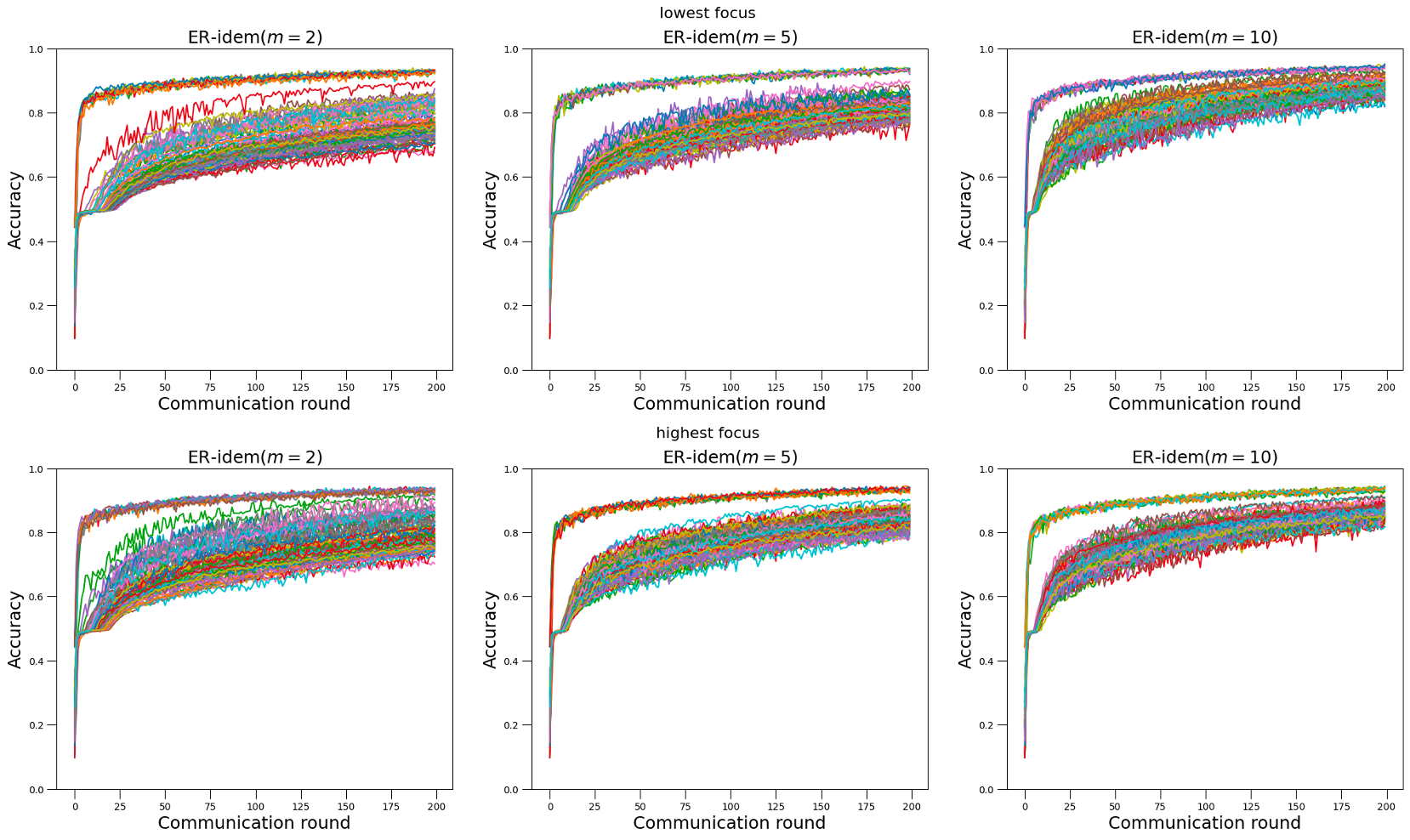}
    \caption{Accuracy over time in ER networks (all nodes, G2 data assigned based on clustering coefficient). From left to right: increasing values of connectedness; from top to bottom: lowest-focused and highest-focused scenario.}
    \label{fig:er_clust_all}
    \end{minipage}
%
    \begin{minipage}{\textwidth}
    \centering
    \includegraphics[width=\textwidth]{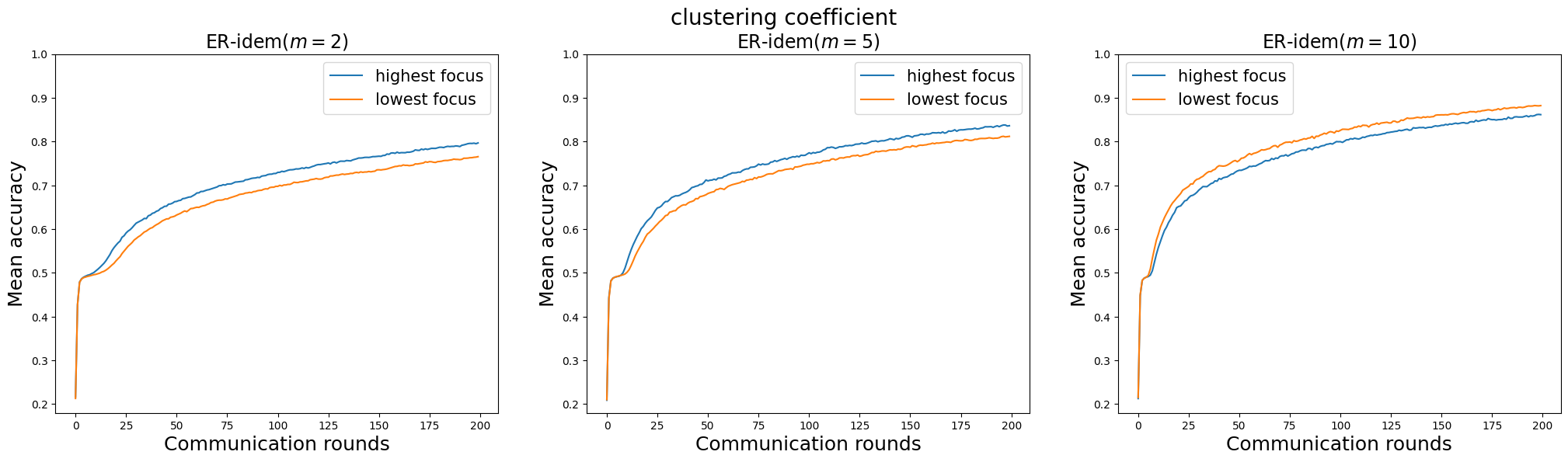}
    \caption{Mean accuracy over time in ER networks (average across G1 nodes only, G2 data assigned based on clustering coefficient). Highest-focused (blue) and lowest-focused (orange) cases.}
    \label{fig:er_clust_high_vs_low}
    \end{minipage}
%
    \begin{minipage}{\textwidth}
    \centering
    \includegraphics[width=\textwidth]{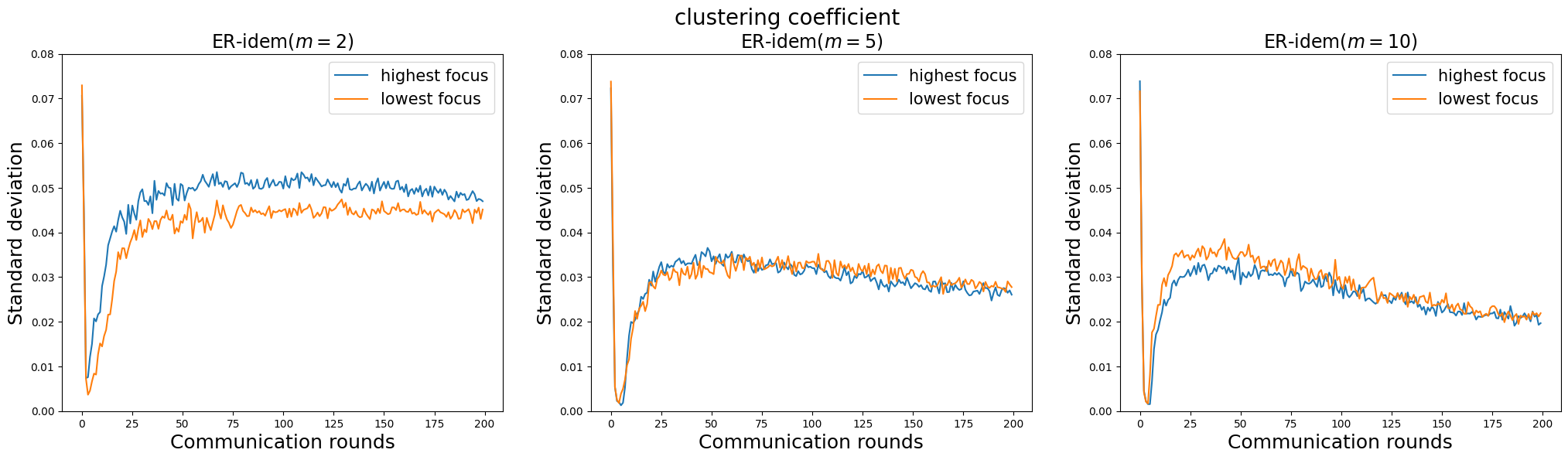}
    \caption{Standard deviation of accuracy over time in ER networks (std dev. across G1 nodes only, G2 data assigned based on clustering coefficient). Left to right: increasing value of the connectedness.}
    \label{fig:er_clust_std_all}
    \end{minipage}
\end{figure}

As we remarked when discussing~\Cref{fig:er_clust_scatterplot}, the clustering coefficient has no significant relation with the degree centrality (and, as a result, with the betweenness centrality) in our ER graphs. This implies that the degree of nodes can be high or low regardless of their clustering coefficient. Given that degree centrality seems to be the main driver of knowledge spreading, we do not expect much variation in the accuracy between the highest and lowest focus cases. This is evident in~\Cref{fig:er_clust_all}. This implies that the \ul{clustering coefficient does not reflect in any meaningful way the model aggregation and dissemination process. For this reason, whether G2 data are assigned to nodes with low or high clustering coefficients does not make a difference.}
The similar performance between the highest and lowest focus cases is further confirmed by the mean accuracy curves of all G1 nodes showed in \Cref{fig:er_clust_high_vs_low}, where we can see that the curves have a small gap, with the highest focus case being above the lowest focus one with the exception of the last scenario. 
In the high-focus case an increase of connectivity does not help much since high-focus nodes are anyway poorly connected. In the low-focus case this is the opposite,  and rightmost plot of \Cref{fig:er_clust_high_vs_low} shows that, in this case, the effect of higher connectivity for the low-focus case is more important then allocating G2 on most central nodes. 

In \Cref{fig:er_clust_std_all} we show the standard deviation of the accuracy for all the scenarios and all networks. As we can see, the curves have all the same trend and are closer together. Therefore, for better visualization, we plot the different network realizations separately. For the ER-idem($m=2$) case, the highest-focus case is above the lowest-focus counterpart, meaning that the highest-focus shows a higher standard deviation. This is due to the locality of information dissemination given that G2 nodes are well-connected inside their neighbourhood but might not be as homogeneously connected to the rest of the network. 



\subsection{Decentralised learning over Barabási-Albert graphs}
\label{sec:results_ba}

\begin{figure}[t!]
    \begin{minipage}{\textwidth}
    \includegraphics[width=\textwidth]{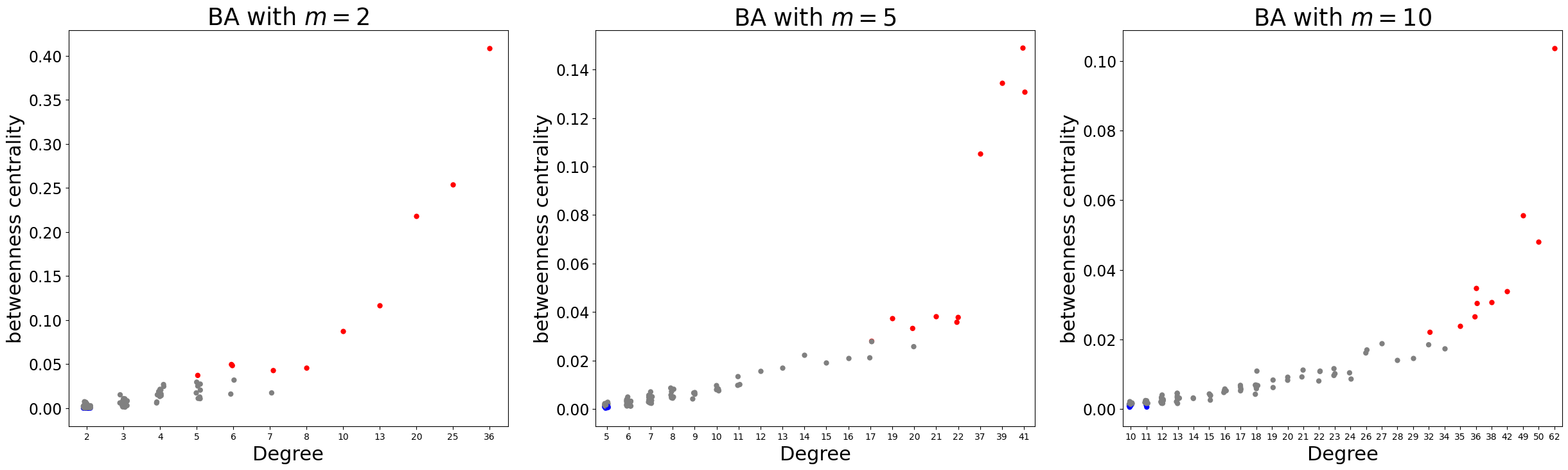}
    \caption{Scatterplot of the BA graphs. Betwenness centrality vs degree. Red dots:
highest-focus nodes; blue dots: lowest-focus nodes.}
    \label{fig:ba_bet_scatterplot}      
    \end{minipage}
    \begin{minipage}{\textwidth}
    \includegraphics[width=\textwidth]{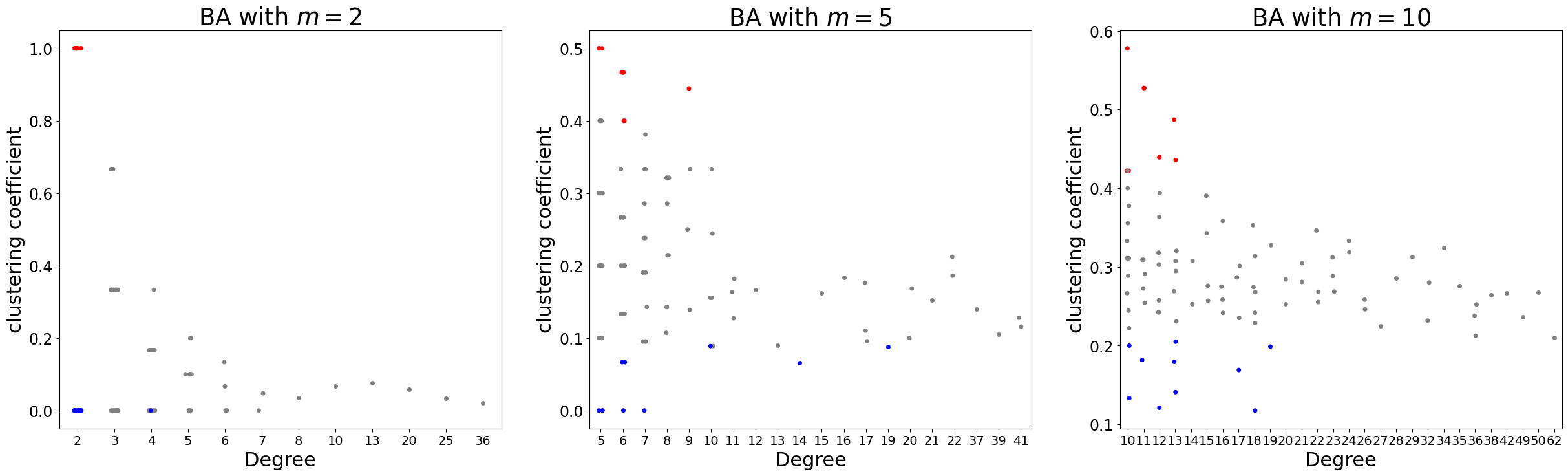}
    \caption{Scatterplot of the BA graphs. Clustering coefficient vs degree. Red dots:
highest-focus nodes; blue dots: lowest-focus nodes.}
    \label{fig:ba_clust_scatterplot}      
    \end{minipage}
\end{figure}

In this section, we focus our analysis on the Barabási-Albert topology, with three different settings for the parameter related to the preferential attachment: $m \, = \,2,5,10$. Preliminarily, we investigate, as we did for ER, the relationship between the different centrality measures used for G2 data assignment (\Cref{fig:ba_bet_scatterplot}). The relationship between degree and betweenness centrality is still one of direct proportionality, as expected, but with a more pronounced quadratic trend. Note, also, that nodes reach much higher values for both centrality metrics with BA, which follows directly from the topological properties of BA networks. Moreover, due to the power law nature of the BA degree distribution, 
%
the nodes chosen for the highest focus case (those indicated by the red dots) span a wider range of degrees than in ER. 
When looking at the degree centrality vs clustering coefficient (\Cref{fig:ba_clust_scatterplot}), BA graphs exhibit nodes with both high and low clustering coefficients concentrated on the left-hand side of the scatterplot, suggesting a generally low degree for these nodes (with the exception of some nodes with a medium-range degree when $m=5$). 

\subsubsection{BA: Degree centrality used for G2 data assignment}
\label{sec:results_ba_degree}

\begin{figure*}[p]
    \begin{minipage}[t]{\textwidth}
    \centering
    \includegraphics[width =1\textwidth]{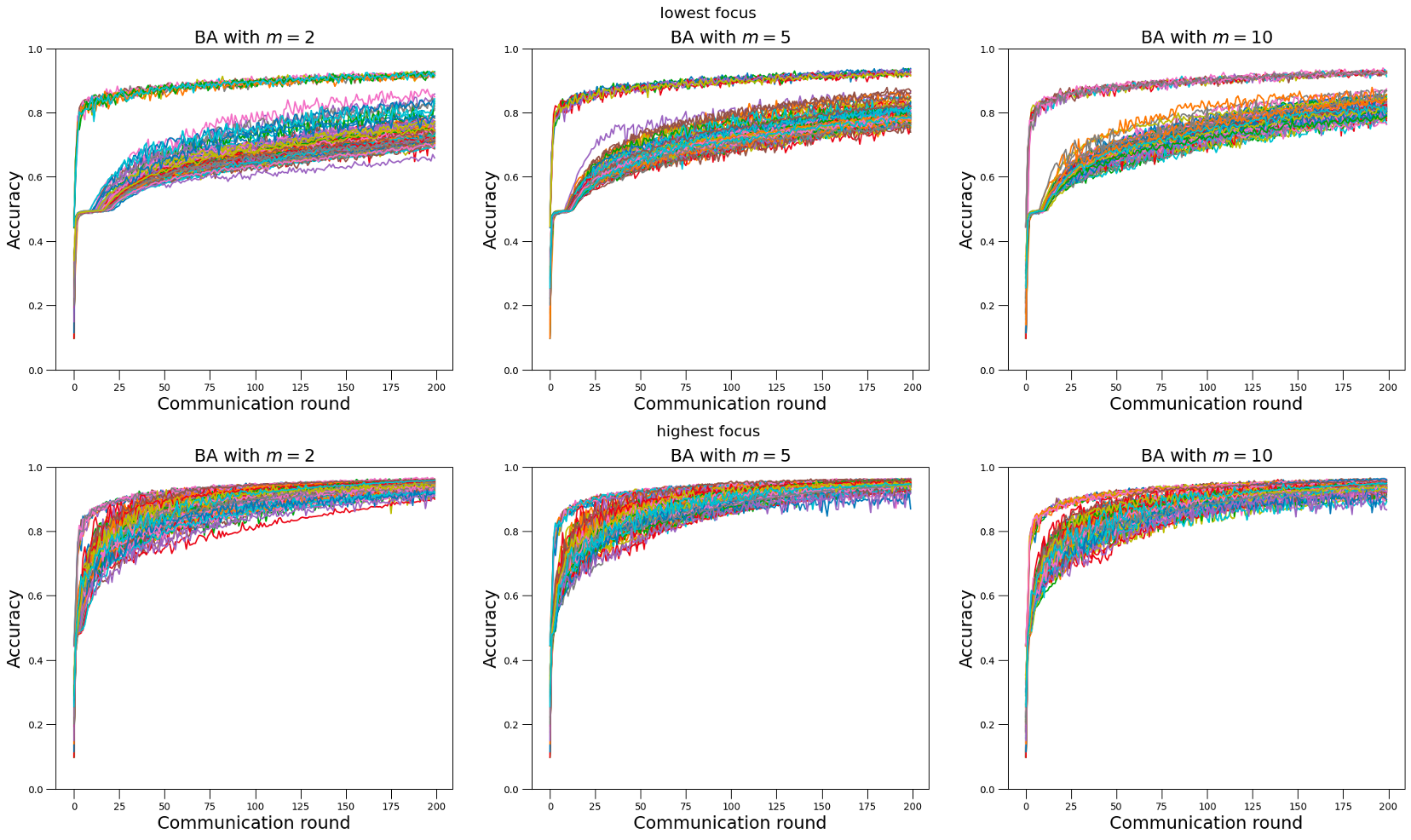}
    \caption{Accuracy in BA networks (G1 nodes, G2 data assigned based on degree centrality). From left to right, the parameter $m$ increases; from top to bottom: edge-focused and hub-focused scenario.} 
    \label{fig:ba_accuracy_tot}
    \end{minipage}
%
 \begin{minipage}[t]{\textwidth}
    \centering
    \includegraphics[width=1\textwidth]{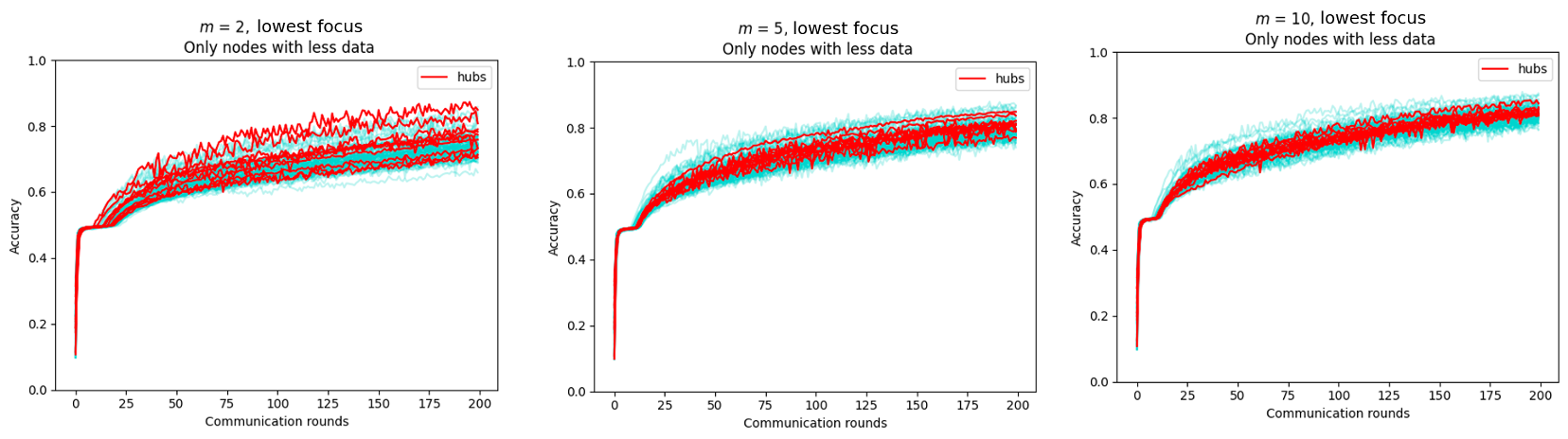} 
    \caption{Accuracy in BA networks (nodes with only G1 images, edge-focused scenario, G2 data assigned based on degree centrality), high-degree nodes in red.} 
    \label{fig:ba_accuracy_lessdata_hubshighlight}
    \end{minipage}
\begin{minipage}[t]{\textwidth}
    \centering
    \includegraphics[width=0.8\textwidth]{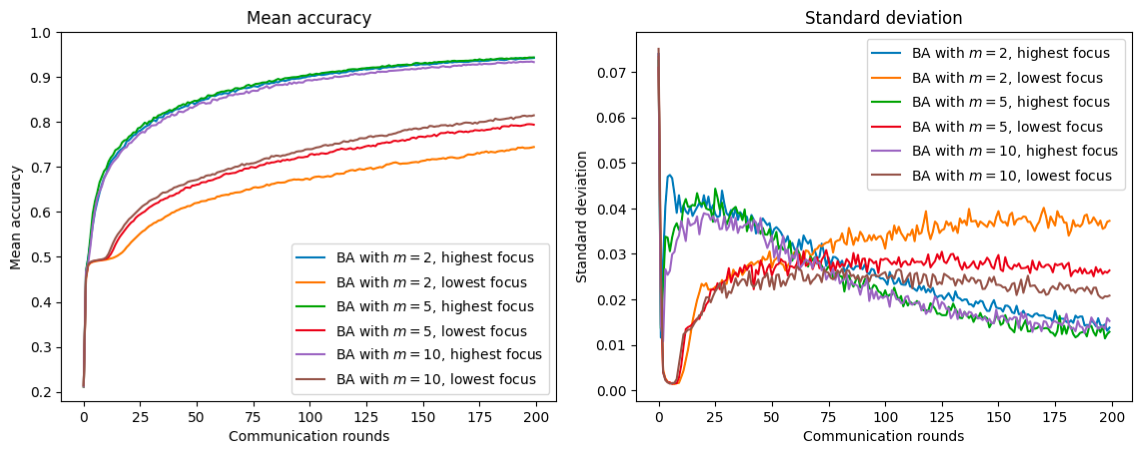}
    \caption{BA (G1 nodes, G2 data assigned based on degree centrality): accuracy mean and std.} 
    \label{fig:ba_mean_acc_std_all}
    \end{minipage}
\end{figure*}

We show the accuracy over time for the case in which degree centrality is used to drive the G2 data assignment in Figure~\ref{fig:ba_accuracy_tot}. First, in the hub-focused case (bottom row), the performance for varying values of the minimum degree $m$ is basically indistinguishable (this is confirmed by looking at the average and standard deviation of the accuracy in Figure~\ref{fig:ba_mean_acc_std_all}). This means that hubs (in this case, high-degree nodes are \emph{real} hubs) spread knowledge extremely efficiently, irrespective of the connectivity of the rest of the nodes.
%
%
The edge-focused case (top row of Figure~\ref{fig:ba_accuracy_tot}) is more challenging. As in the case of ER networks, edge nodes are unable to spread their knowledge efficiently and the accuracy gap between edge nodes and non-edge nodes remains strong all throughout. In Figure~\ref{fig:ba_mean_acc_std_all}, we observe that larger values of $m$ (i.e., stronger connectivity) help improve the average accuracy in the edge-focused case but not significantly, and the variability is reduced (as shown by the standard deviation curves on the right-hand side plot). 

Given the distinctive features of high-degree nodes in BA network, we ask whether they are at an advantage or disadvantage in an edge-focused scenario. Thus, in Figure~\ref{fig:ba_accuracy_lessdata_hubshighlight} we focus on the G1 nodes for the edge-focused scenario, highlighting in red the behaviour of high degree nodes. For smaller $m$, the neighbourhood of these hubs is smaller, hence the aggregation is more influenced by the capabilities of the specific nodes composing the neighbourhood, resulting in higher variability. As $m$ increases, the neighbourhood of hubs becomes so large that it will include both ``good'' models (with G1 and G2 knowledge) and models knowing only about G1. The ``good'' models are then more likely to get lost in the average, hence the accuracy of previously well-performing hubs goes down as $m$ increases. However, the overall accuracy for G1 nodes does not decrease. This is because G1 nodes that are not hubs increase their connectivity as $m$ increases. This results in better average accuracy (some hubs get worse, but other G1 nodes get better) and reduced variability (Figure~\ref{fig:ba_mean_acc_std_all}).

\subsubsection{BA: Betweenness centrality used for G2 data assignment}
\label{sec:results_betweenness}

\begin{figure*}[t!]
    \begin{minipage}[t]{\textwidth}
    \centering
    \includegraphics[width=\textwidth]{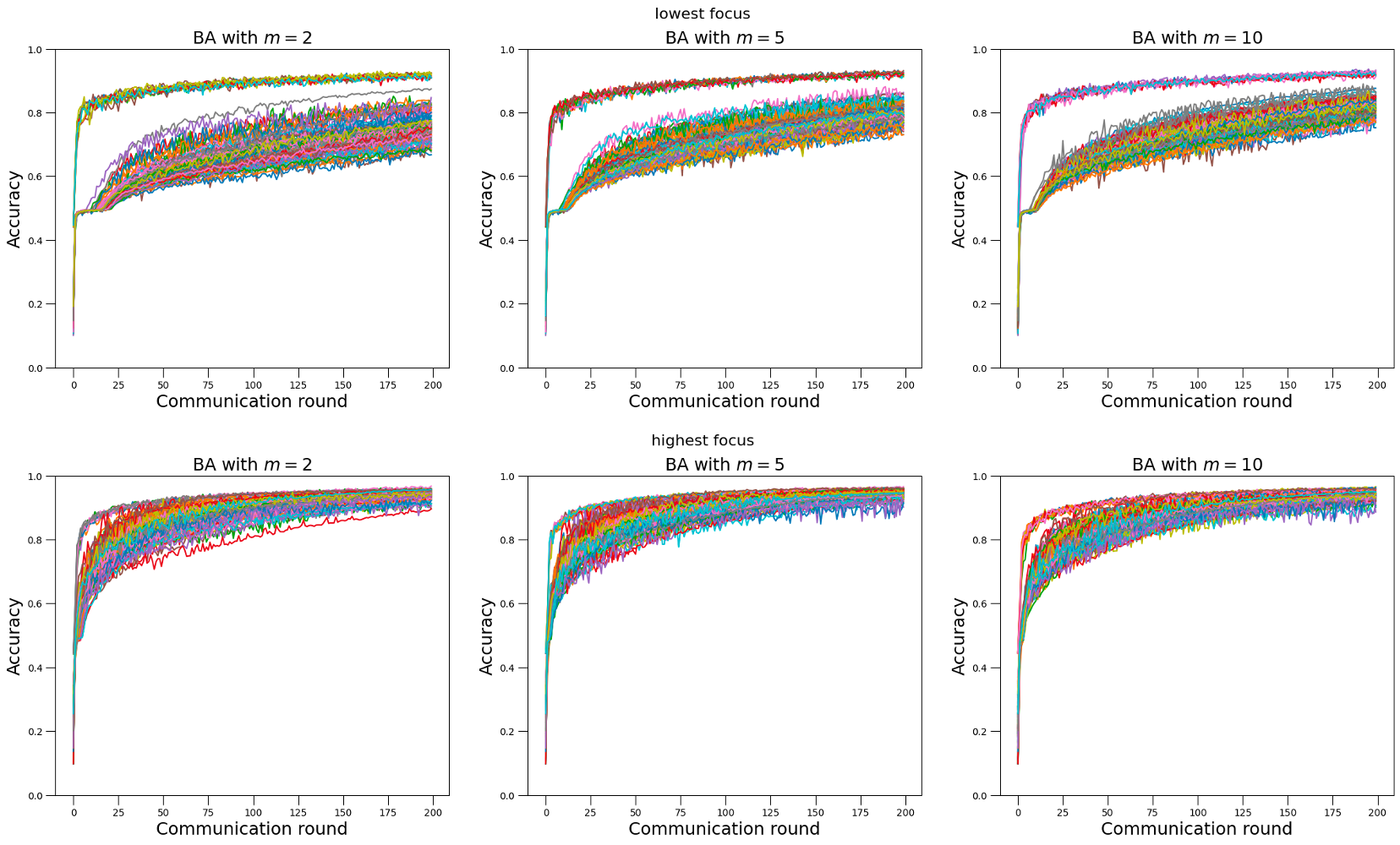}
    \caption{Accuracy in BA networks (all nodes, G2 data assigned based on betweenness centrality). From left to right, increasing values of $m$; from top to bottom: lowest-focused and highest-focused scenario.}
    \label{fig:ba_bet_all}
    \end{minipage}
%
    \begin{minipage}[t]{\textwidth}
    \centering
    \includegraphics[width=0.9\textwidth]{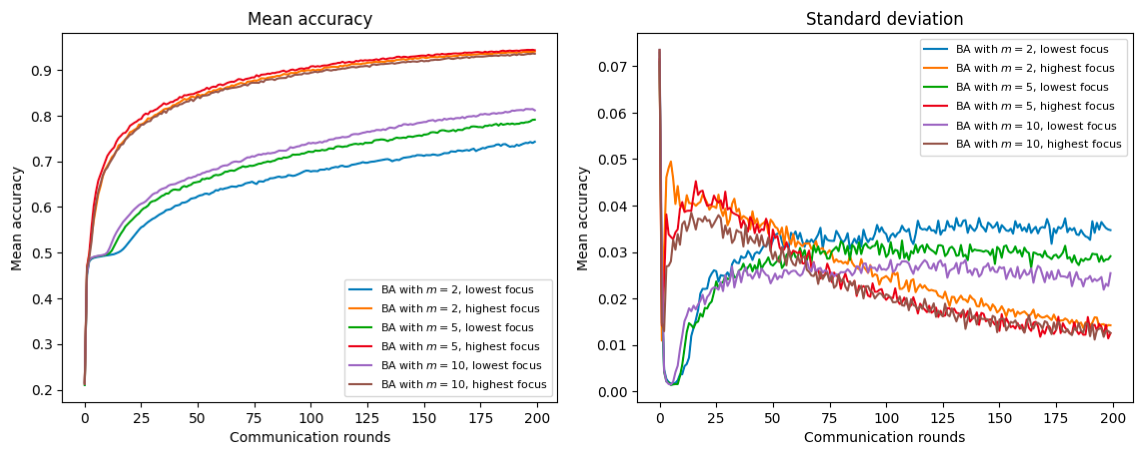}
    \caption{BA (G1 nodes), G2 data assigned based on betweenness centrality): accuracy mean and std for betweenness centrality.}
    \label{fig:ba_bet_mean_acc_std_all}
    \end{minipage}
\end{figure*}

Since the betweenness centrality is proportional to the degree (\Cref{fig:ba_bet_scatterplot}), we expect the G2 data assignment based on betweenness centrality to perform similarly to its degree-based counterpart.
In \Cref{fig:ba_bet_all}, in the lowest focus case (top row), we see an increasingly narrow curve meaning that as we increase the connectedness, the knowledge is more easily spread and this results in less variation in the performances. The overall behaviour is almost identical to the degree-based one, with almost no difference in the performances in the highest focus case.
In \Cref{fig:ba_bet_mean_acc_std_all}, we show the mean accuracy and the standard deviation for each network and case. Again, the same behavior as in the degree-based case is observed.

\subsubsection{BA: Clustering coefficient used for G2 data assignment}
\label{sec:results_ba_clustering}

\begin{figure}[p!]
    \begin{minipage}[t]{\textwidth}
    \centering
    \includegraphics[width=\textwidth]{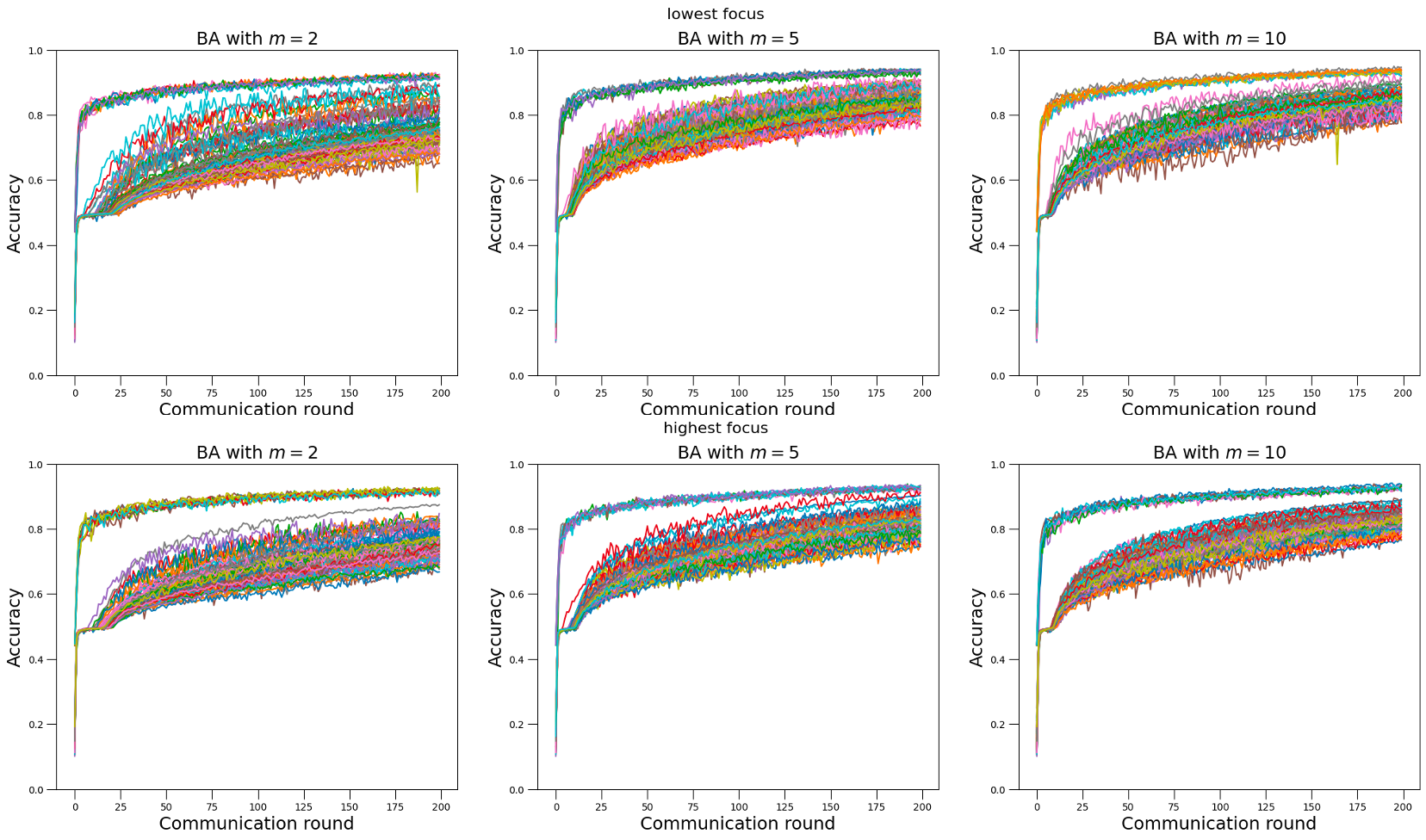}
    \caption{Accuracy in BA networks (all nodes, G2 data assigned based on clustering coefficient). From left to right, increasing values of $m$; from top to bottom: lowest-focused and highest-focused scenario.}
    \label{fig:ba_clust_all}
    \end{minipage}
    \begin{minipage}[t]{\textwidth}
    \centering
    \includegraphics[width=\textwidth]{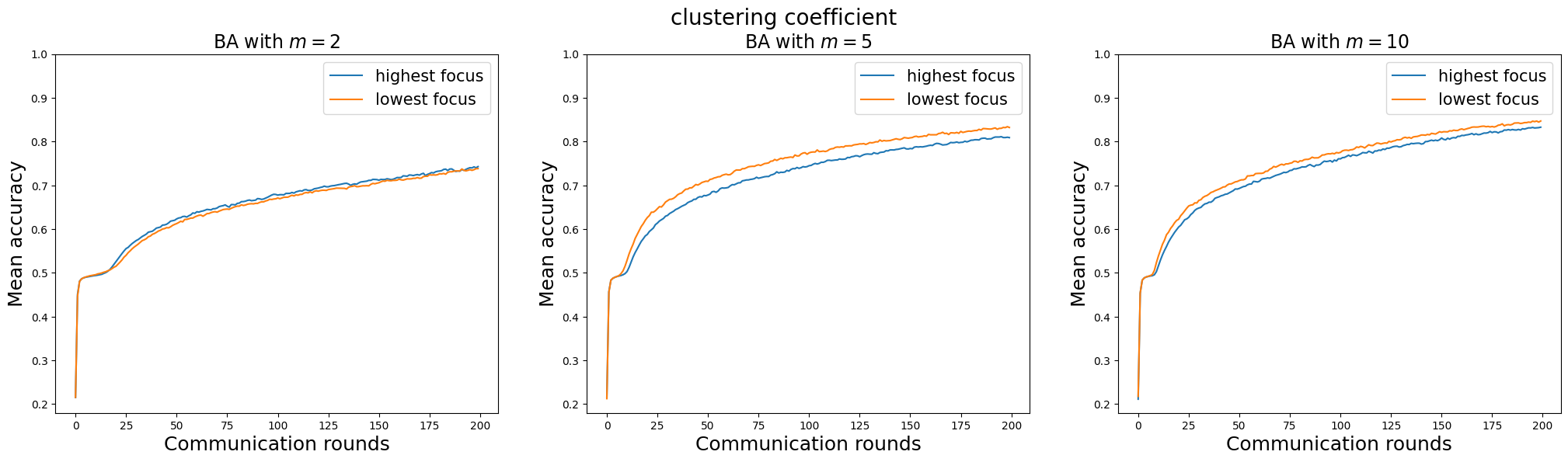}
    \caption{Accuracy in BA networks (average across all nodes, G2 data assigned based on clustering coefficient). From left to right, increasing values of $m$; from top to bottom: lowest-focused and highest-focused scenario.}
    \label{fig:ba_clust_high_vs_low}
    \end{minipage}
%
    \begin{minipage}[t]{\textwidth}
    \centering
    \includegraphics[width=\textwidth]{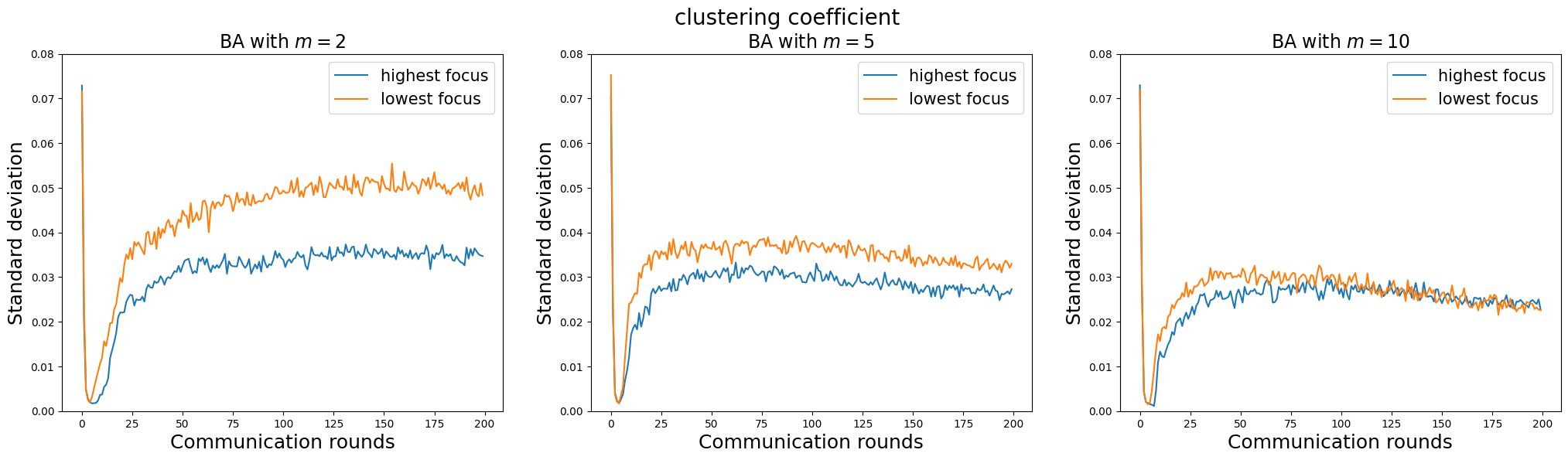}
    \caption{BA (G1 nodes): standard deviation, G2 data assigned based on clustering coefficient.}
    \label{fig:ba_clust_std_all}
    \end{minipage}
\end{figure}

Looking at the individual curves of the accuracy over time in \Cref{fig:ba_clust_all}, we can see that the behaviour is similar between the highest and lowest focus cases, as a result of the negligible difference among the degrees of the selected nodes for G2 assignment (see \Cref{fig:ba_clust_scatterplot}). However, as expected from the scatterplot in \Cref{fig:ba_clust_scatterplot}, the lowest focus case shows a higher variability between the performances of the different nodes. As such, some nodes are able to reach higher values of accuracy (see case $m=2$). The reasoning is the same as for the ER networks: the nodes with high clustering coefficient tend to keep the information locally, whereas the ones with low clustering coefficient spread it more broadly if they have also a high degree. This spreading results in higher variability.
Figures~\ref{fig:ba_clust_high_vs_low} and~\Cref{fig:ba_clust_std_all} simply confirm this trend. As for the ER, also in~\Cref{fig:ba_clust_std_all} the curves are closer together, thus we show the network realizations separately.

\subsection{ER vs BA comparison}
\label{sec:er_vs_ba}

As explained in \Cref{sec:network-settings}, we have chosen our ER and BA settings such that the corresponding graphs are idempotent. This allows us to compare directly and fairly the impact of the network topology on the learning process for the same value $m$. For clarity, we focus on the mean accuracy, so that the plot is not overclutted with curves and a direct comparison is more immediate.

\subsubsection{Degree centrality}

\begin{figure}[t!]
    \begin{minipage}[t]{\textwidth}
    \centering
    \includegraphics[width=\textwidth]{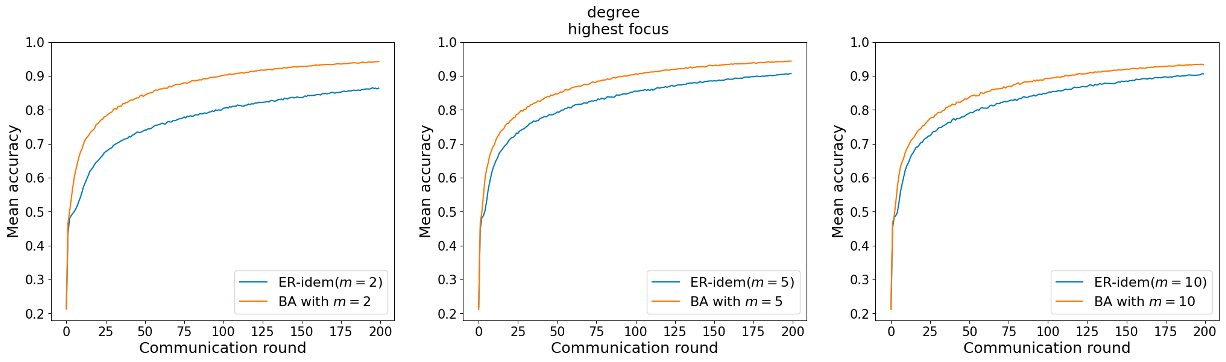}
    \caption{Mean accuracy of the Barabasi Albert (orange line) and the Erdős-Rényi counterpart (blue line) in the hub-focused case. Going from left to right: $m=2,5,10$. G2 data assigned based on degree centrality.}
    \label{fig:er_vs_ba_hubs_focused_degree}
    \end{minipage}
    \begin{minipage}[t]{\textwidth}
    \centering
    \includegraphics[width=\textwidth]{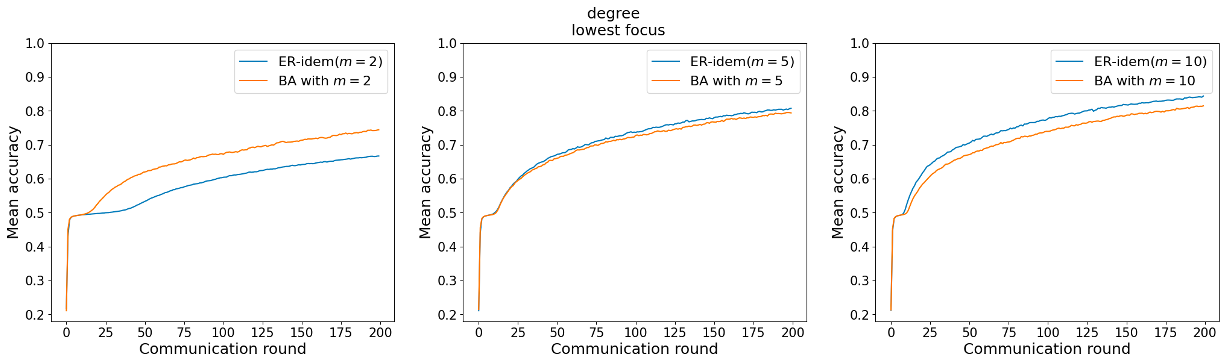}
    \caption{Mean accuracy of the Barabasi Albert (orange line) and the Erdős-Rényi counterpart (blue line) in the edge-focused case. Going from left to right: $m=2,5,10$. G2 data assigned based on degree centrality.}
    \label{fig:er_vs_ba_edges_focused_degree}
    \end{minipage}
\end{figure}

In the hub-focused case, as shown in \Cref{fig:er_vs_ba_hubs_focused_degree}, the Barabasi Albert performs better than the ER counterpart. We can see that the difference decreases as we increase the number of connections in the graphs (i.e. going from left to right). This is easily explained by looking at the characteristics of the topology of the two types of graphs. The Barabási-Albert graph, in fact, is a graph that exhibits a heavy tail in the degree distribution, meaning that it has few nodes with high degree and many nodes with low degree. This means that the high degree nodes are well connected in the graph and can exert their influence much more efficiently than the Erdős-Rényi counterpart. On the other hand, the Erdős-Rényi graph is a random graph, thus the nodes with high degree are not necessarily well connected inside the graph. However, this difference gets smaller as we increase the mean number of connections: all nodes in general enjoy a better connectivity, hence the network can compensate for the lack of strong hubs.

The edge-focused case is more diverse. Here, BA performs better than the ER counterpart only for $m=2$. In the other two cases, ER is the one with better performance, as reported in \Cref{fig:er_vs_ba_edges_focused_degree}. 
This can be confirmed by looking at the algebraic connectivity of the graphs (\Cref{tab:algebraic_connectivity}). Algebraic connectivity is a measure of the connectedness of a graph. A graph with high algebraic connectivity is well-connected and can quickly spread information, while a graph with low algebraic connectivity can only spread information slowly. The higher the algebraic connectivity, the higher the number of paths for information to travel through the graph. As shown in \Cref{tab:algebraic_connectivity}, the algebraic connectivity of the $m=2$ case is lower for the Erdős-Rényi with respect to the Barabási-Albert. 

\begin{table}[t!]
\scriptsize
\centering
\begin{tabular}{@{}cc cc cc @{}}
\toprule
\multicolumn{2}{c}{$m=2$}   & \multicolumn{2}{c}{$m=5$}    & \multicolumn{2}{c}{$m=10$}   \\ 
\cmidrule(lr){1-2} \cmidrule(lr){3-4} \cmidrule(lr){5-6}
BA & ER & BA & ER & BA & ER \\
$0.612$         & $0.395$     & $2.989$         & $2.998$     & $4.718$         & $8.143$     \\ \bottomrule
\end{tabular}
\caption{Table showing the algebraic connectivity values for the BA graphs and their ER counterpart for different values of $m$.}
\label{tab:algebraic_connectivity}
\end{table}

\subsubsection{Betweenness centrality}
\begin{figure}[t!]
    \centering
    \includegraphics[width=\textwidth]{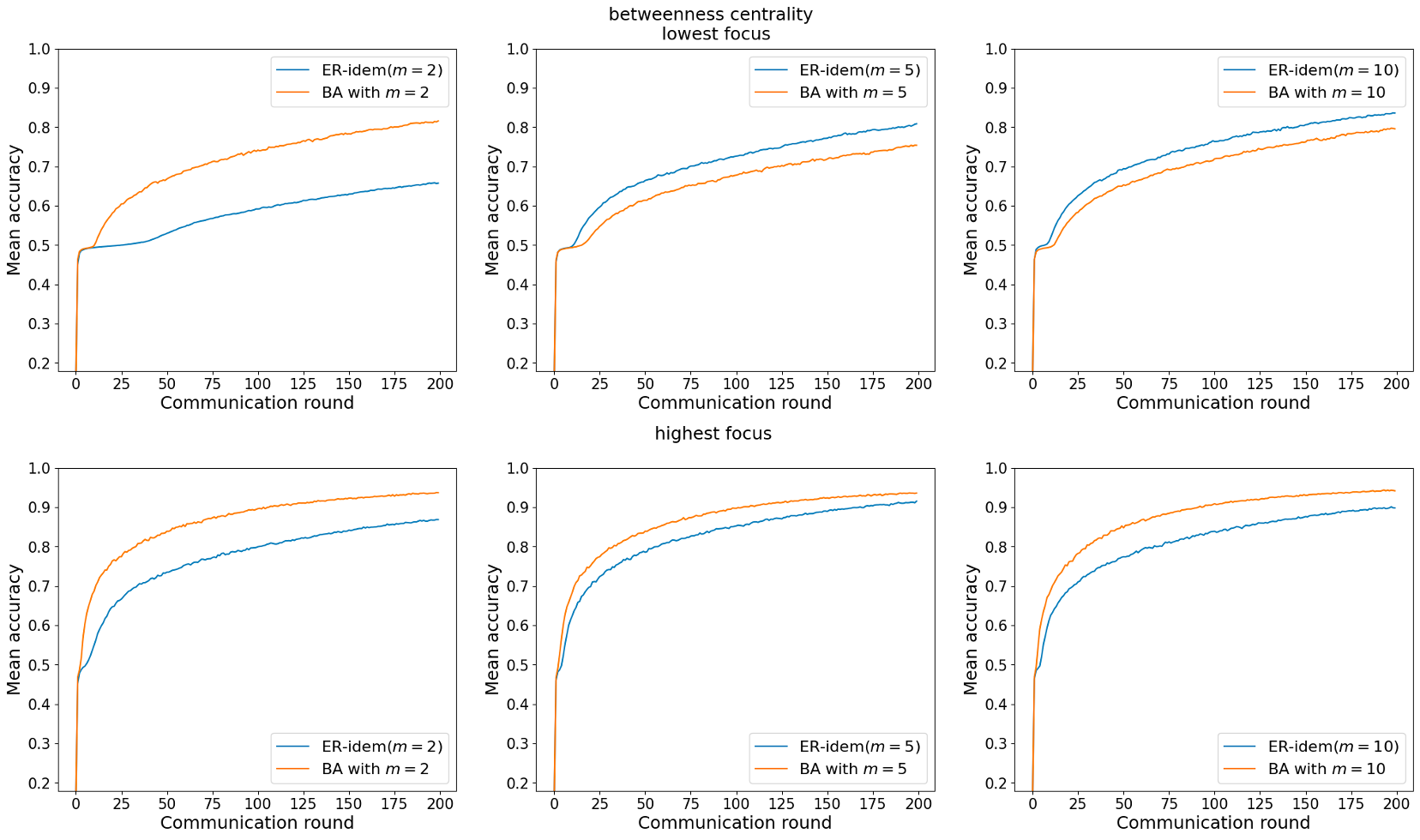}
    \caption{Comparison of mean accuracy ER vs BA. Top to bottom: lowest-focus and highest-focus cases. Left to right: increasing $m$. G2 data assigned based on betweenness centrality.}
    \label{fig:er_vs_ba_betwenneess}
\end{figure}

As already discussed in the previous sections, the betweenness centrality shows both for the ER and the BA networks a proportional relationship to the degree. Therefore, we expect to have similar results for the betweenness centrality.
Similarly to the degree case, in the lowest focus, the BA performs better than the ER counterpart only for $m=2$, as shown in \Cref{fig:er_vs_ba_betwenneess}. This means that, comparing these results with Figure~\ref{fig:er_vs_ba_edges_focused_degree}, G2 nodes are not able to sufficiently influence the hubs.

\subsubsection{Clustering coefficient}
\begin{figure}[t!]
    \centering
\includegraphics[width=\textwidth,keepaspectratio]{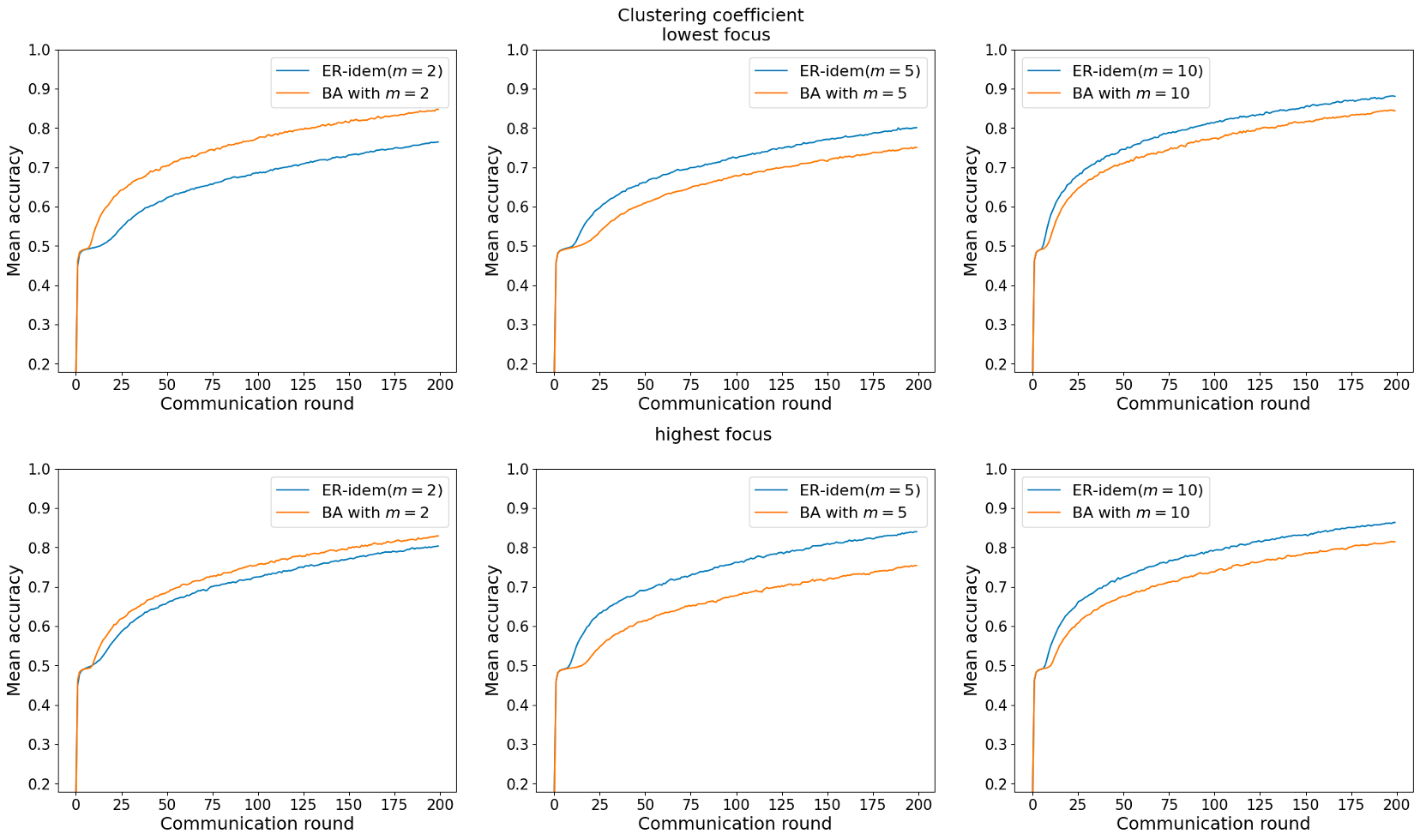}
    \caption{Comparison of mean accuracy ER vs BA. Left to right: increasing $m$. Top to bottom: lowest-focus and highest-focus. G2 data assigned based on clustering coefficient.}
    \label{fig:er_vs_ba_clustering}
\end{figure}

Similarly as above, also the clustering coefficient shows for the ER-idem($m=2$) a worse performance than the BA, see \Cref{fig:er_vs_ba_clustering}. Again, this is due to the algebraic connectivity that helps the knowledge spreading, as discussed for Figure~\ref{fig:er_vs_ba_betwenneess} and earlier.

\subsection{Decentralised learning over Stochastic Block Model graphs}
\label{sec:results_sbm}

\begin{figure}[p!]
    \begin{minipage}{\textwidth}
    \centering
    \includegraphics[width =\textwidth]{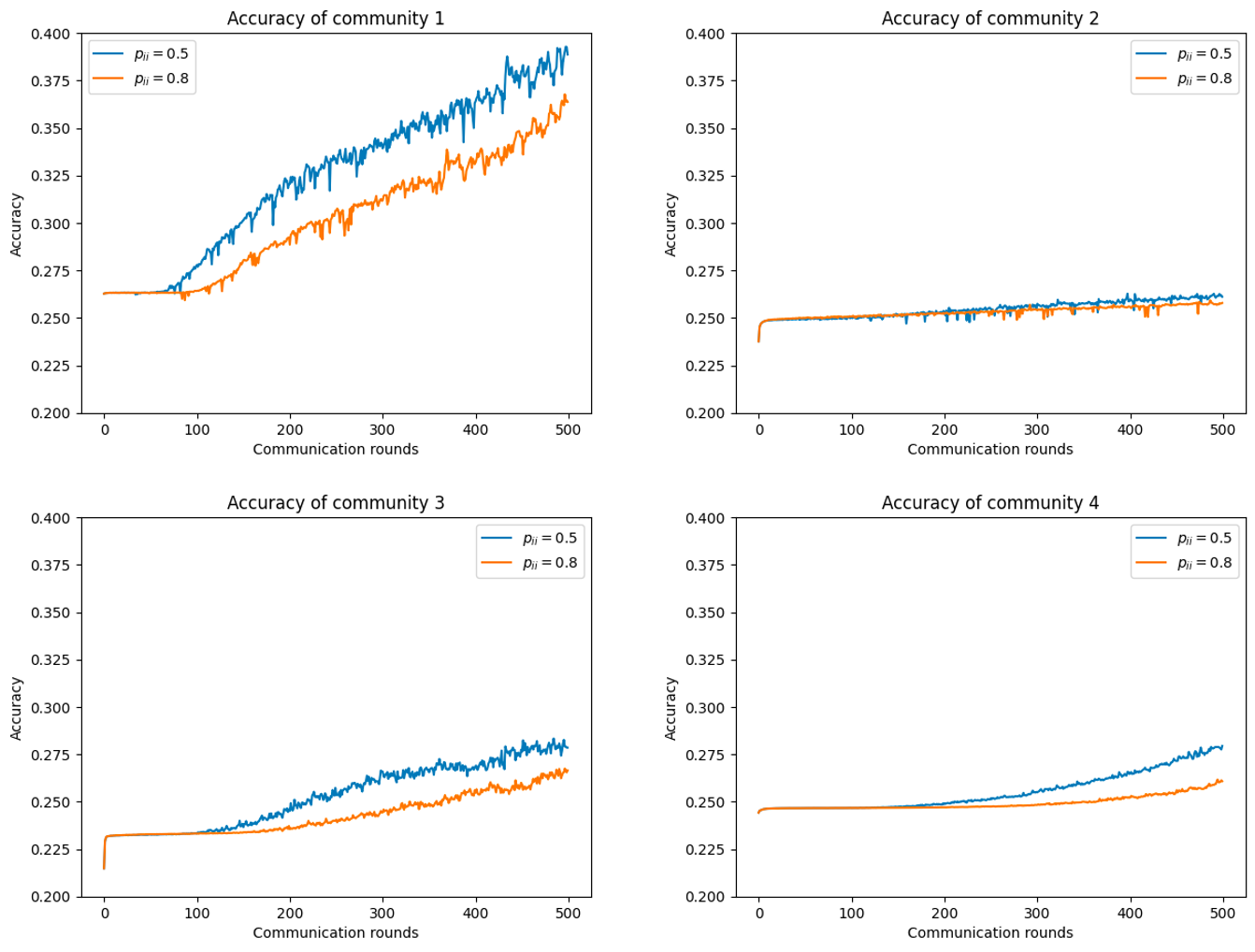}
    \caption{Mean accuracy over time in SBM communities.} 
    \label{fig:sbm_mean_acc_per_community}
    \end{minipage} \vspace{1cm}
    \begin{minipage}{\textwidth}
    \scriptsize
    \centering
    \begin{tabular}{@{}lllll@{}}
    \toprule
       \textbf{Class} & \textbf{$C_1$} & \textbf{$C_2$} & \textbf{$C_3$} & \textbf{$C_4$} \\
       & (-, 5,9,7) & (5,-,7,3) & (9,7,-,8) & (7,3,8,-)\\\midrule
       0 & \textbf{0.9961} & 0.0002 & 0.0004 & 0.0043 \\ 
       1 & \textbf{0.9992} & 4e-05 & 0.0684 & 0.0079 \\ \hline
       2 & 0.0 & \textbf{0.9868} & 4e-05 & 0 \\ 
       3 & 0.0146 & \textbf{0.9802} & 0 & 0.0002 \\ \hline
       4 & 0.1824 & 0.0076 & \textbf{0.9971} & 0.0006 \\ 
       5 & 0.0011 & 4e-05 & \textbf{0.9972} & 0.0011 \\ \hline
       6 & 0.0039 & 0 & 0.0003 & \textbf{0.9979} \\
       7 & 0.272 & 0.0101 & 0.0225 & \textbf{0.9966} \\ \bottomrule \\
   \end{tabular} 
   \caption{Accuracy per MNIST class and community for SBM with $p_{ii} \, = \, 0.5$. For each community, we report in square brackets the number of edges pointing toward external community 1, 2, 3, 4, respectively.} 
    \label{tab:confmat_05}
   \end{minipage} 
   \end{figure}

The SBM topology is different from the previous two as it features four clearly separated communities, with sporadic intercommunity links. For the intracommunity connectivity, we test two scenarios (lower and higher intracommunity connectivity, corresponding to $p_{ii}= 0.5$ and $p_{ii}= 0.8$). Recall that each community holds two non-overlapping MNIST classes (hence, classes 8 and 9 are discarded). Using only intracommunity information, nodes can at most achieve a 0.25 accuracy (perfect classification of the two classes in their training data, zero knowledge on the other six). In order to go beyond 0.25, knowledge must be circulated across communities. In these settings we are interested in analysing (i) whether the sporadic edges between communities are sufficient for exchanging the knowledge built intra-community and (ii) to what extent the external knowledge permeates through the communities.   
Figure~\ref{fig:sbm_mean_acc_per_community} shows that the latter is happening: with $p_{ii}=0.5$ (less dense communities, blue lines) the average accuracy grows faster than with $p_{ii}=0.8$. 
The figure also reveals the presence of stragglers, for whom catching up with the rest of the network takes some time. Interestingly, entire communities appear to be stragglers. Table~\ref{tab:confmat_05} shows that communities are, as expected, very good at classifying classes they have in their local training data. Vice versa, it is very hard for external knowledge to enter the communities. Table~\ref{tab:confmat_05} also shows the number of links pointing towards external communities, which are the conduit for knowledge diffusion. Community $C_2$ enjoys fewer external links, and indeed its learning process is very slow and mediocre (Figure~\ref{fig:sbm_mean_acc_per_community}). However, Community $C_1$ has only slightly fewer links than Community $C_3$, but its learning process is faster and more accurate. This observation suggests that the specific class distribution assigned to each community might play a crucial role in determining its learning performance. 

\begin{table}[h]
    \centering
    \begin{tabular}{c}
    \toprule
        Source $\rightarrow$ Destination  \\
        \midrule
         $C_1$ $\rightarrow$ $C_2$ \\
         $C_2$ $\rightarrow$ $C_3$ \\
         $C_3$ $\rightarrow$ $C_4$ \\
         $C_4$ $\rightarrow$ $C_1$ \\
        \bottomrule
    \end{tabular}
    \caption{Data swap between communities.}
    \label{tab:swap}
\end{table}

\begin{figure}[p!]
\begin{minipage}{\textwidth}
        \includegraphics[width =\textwidth]{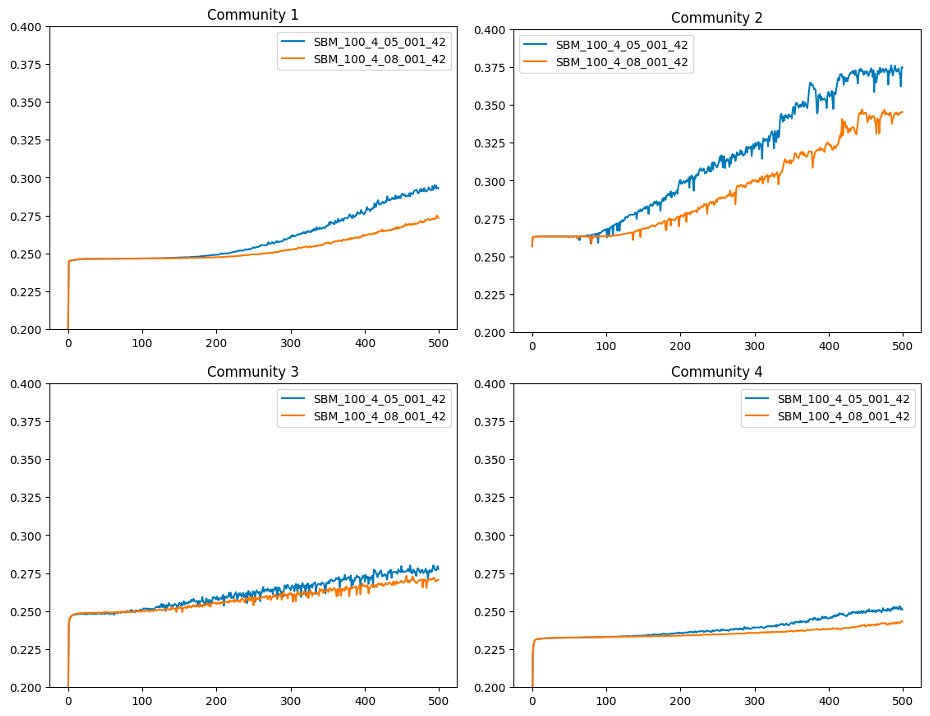}
    \caption{Mean accuracy over time in SBM communities, with swapped datasets.}
    \label{fig:sbm_accuracy_swapped}
\end{minipage}\hspace{1cm}
\begin{minipage}{\textwidth}
    \vspace{6pt}
    \includegraphics[width =\textwidth]{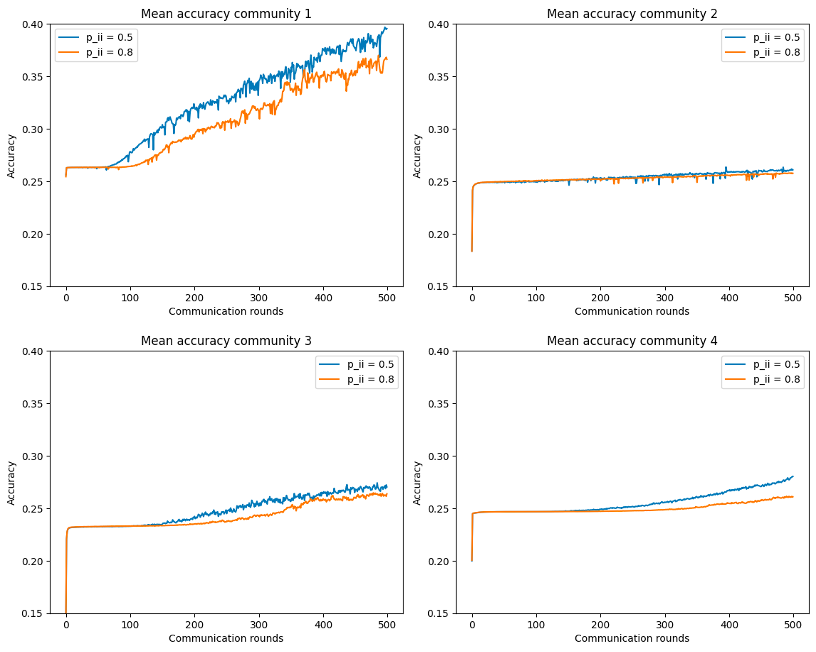}
    \caption{Mean accuracy in SBM communities, with 5 local training epochs instead of 100.}
    \label{fig:sbm_accuracy_5localep}
\end{minipage}
\end{figure}

To validate the above hypothesis, we swapped the datasets among the four communities as reported in \Cref{tab:swap}. If the learning performance is highly dependent on the dataset, one would expect that when the dataset of $C_1$ is moved to $C_2$, the improved learning speed and accuracy would move with it. Conversely, if the characteristics of the community itself play a significant role, one might observe different results.
Our findings, shown in \Cref{fig:sbm_accuracy_swapped}, confirmed this assumption, demonstrating that the specific class distribution significantly influences the learning behaviour of each community.

Furthermore, we explored the impact of conducting multiple local training epochs on the dissemination of information within the community.
The objective was to determine whether a high number of local training epochs in between global updates might negatively impact the information diffusion when communities are tightly knitted. To examine this, we conducted a simulation under identical conditions as in the original experiment but with significantly fewer local training epochs per communication round: specifically, we reduced it to 5 local training epochs as opposed to the original 100. The results, presented in \Cref{fig:sbm_accuracy_5localep}, reveal a pattern consistent with the original findings. This suggests that the challenge of information propagation within the community is not inherent to the local training phase but is primarily influenced by structural properties of the network as well as by the correlation between data classes.



%% file: parts_of_main/conclusion.tex
\section{Conclusion}
\label{sec:conclusion}

Fully decentralized learning is becoming increasingly interesting from a research perspective, as it enables the training of AI models at the edge of the Internet. This approach helps alleviate issues related to data and bandwidth burdens on the infrastructure, latency, the necessity for continuous and reliable Internet connectivity, and also addresses principles of data locality and privacy. However, learning in a fully decentralized manner poses significant challenges that researchers have yet to fully address. In this work, we focus on one of these challenges, specifically the interplay between the network structure, on which the learning process evolves, and the final performance of the learning itself. For this purpose, we selected three network topologies, each representing a dominant topological property, and six different methods of distributing training data among nodes. Below, we summarize the main findings of this work.
\begin{description}
\item[Centrality vs neighbourhood density.] Global centrality metrics directly correlate with the achieved performance of decentralized learning. No significant difference between the two global centrality metrics considered (degree and betweenness) is detected for the topologies we have considered. Vice versa, local clustering is a poor predictor of performance, meaning that belonging to tightly vs loosely knit neighbourhoods does not make any direct difference. 
%
\item[Dilution effect.] Exporting knowledge to central nodes is difficult because they perform model aggregation over a large neighbourhood, where a single ``good'' model may get diluted into the neighbourhood average. 
\item[Centrality pull effect.] Central nodes pull the others towards their knowledge (i.e., learned model). The centrality pull effect combined with the dilution effect explains why knowledge spreads easily if it originates on central nodes. 
\item[Degree distribution.] The presence of hubs (originating from the Pareto distribution of the degree in BA networks) is a double-edged sword. Hubs can worsen the dilution effect when knowledge is on peripheral nodes but can boost the learning when knowledge is sourced by central nodes. Also, as a side effect of the dilution problem, the lower the average degree, the easier the knowledge spreading from non-central nodes.
\item[Segregated communities.]  When users are grouped in segregated communities, it is very difficult for knowledge to circulate outside of the community.
\end{description}

\noindent
The above summary illustrates the complex interplay between network topology, decentralized learning, and training data distribution. It conveys how a one-size-fits-all approach is not adequate for decentralized learning. This paper lays the groundwork for further studies, where more complex network topologies and network dynamics can be investigated. Moreover, it highlights the need for more nuanced model aggregation strategies, e.g., for mitigating the dilution effect and addressing the resulting unfair treatment of peripheral nodes.

%% file: parts_of_main/acks.tex
\section*{Acknowledgments}

This work was partially supported by the H2020 HumaneAI Net (952026) and by the CHIST-ERA-19-XAI010 SAI projects. C. Boldrini's and M. Conti's work was partly funded by the PNRR - M4C2 - Investimento 1.3, Partenariato Esteso PE00000013 - "FAIR", A. Passarella's work was partly funded by the PNRR - M4C2 - Investimento 1.3, Partenariato Esteso PE00000001 - "RESTART", both funded by the European Commission under the NextGeneration EU programme.